\documentclass[10pt,twocolumn,letterpaper]{article}

\usepackage{iccv}
\usepackage{times}
\usepackage{epsfig}
\usepackage{graphicx}
\usepackage{amsmath}
\usepackage{amssymb}
\usepackage{xcolor}
\usepackage{booktabs} % To thicken table lines
\usepackage{caption,subcaption}
\usepackage[most]{tcolorbox}
\usepackage{adjustbox}
\usepackage{bbding}
\usepackage{pifont}
\usepackage{wasysym}
\usepackage{amssymb}
\usepackage[breaklinks=true,bookmarks=false]{hyperref}

\iccvfinalcopy % *** Uncomment this line for the final submission

\def\iccvPaperID{3} % *** Enter the ICCV Paper ID here

\ificcvfinal\fi

\newcommand\animagetgt{\adjustbox{valign=m,vspace=0.2pt}{\includegraphics[width=.2\linewidth]{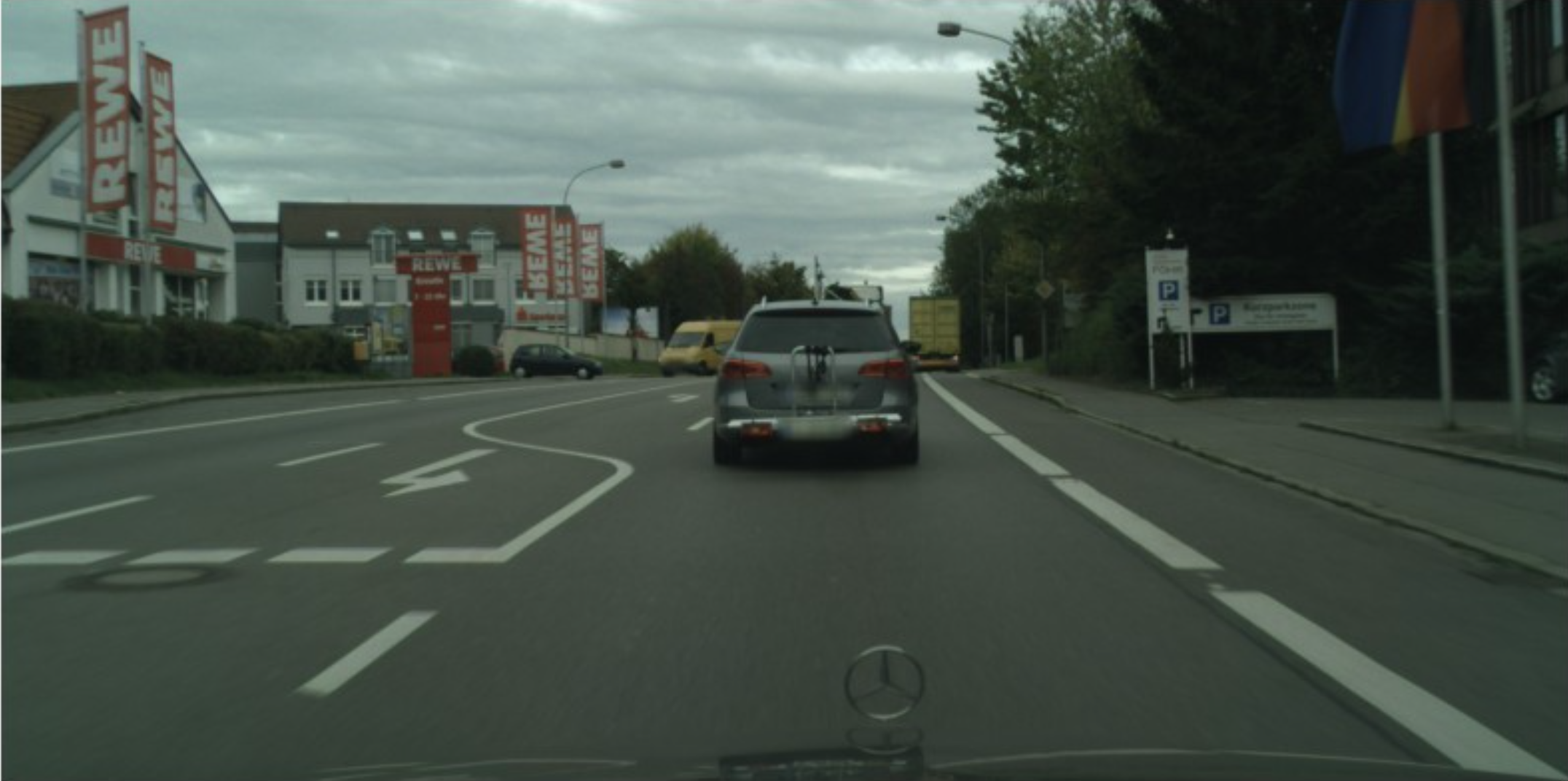}}}
\newcommand\animagegt{\adjustbox{valign=m,vspace=0.2pt}{\includegraphics[width=.2\linewidth]{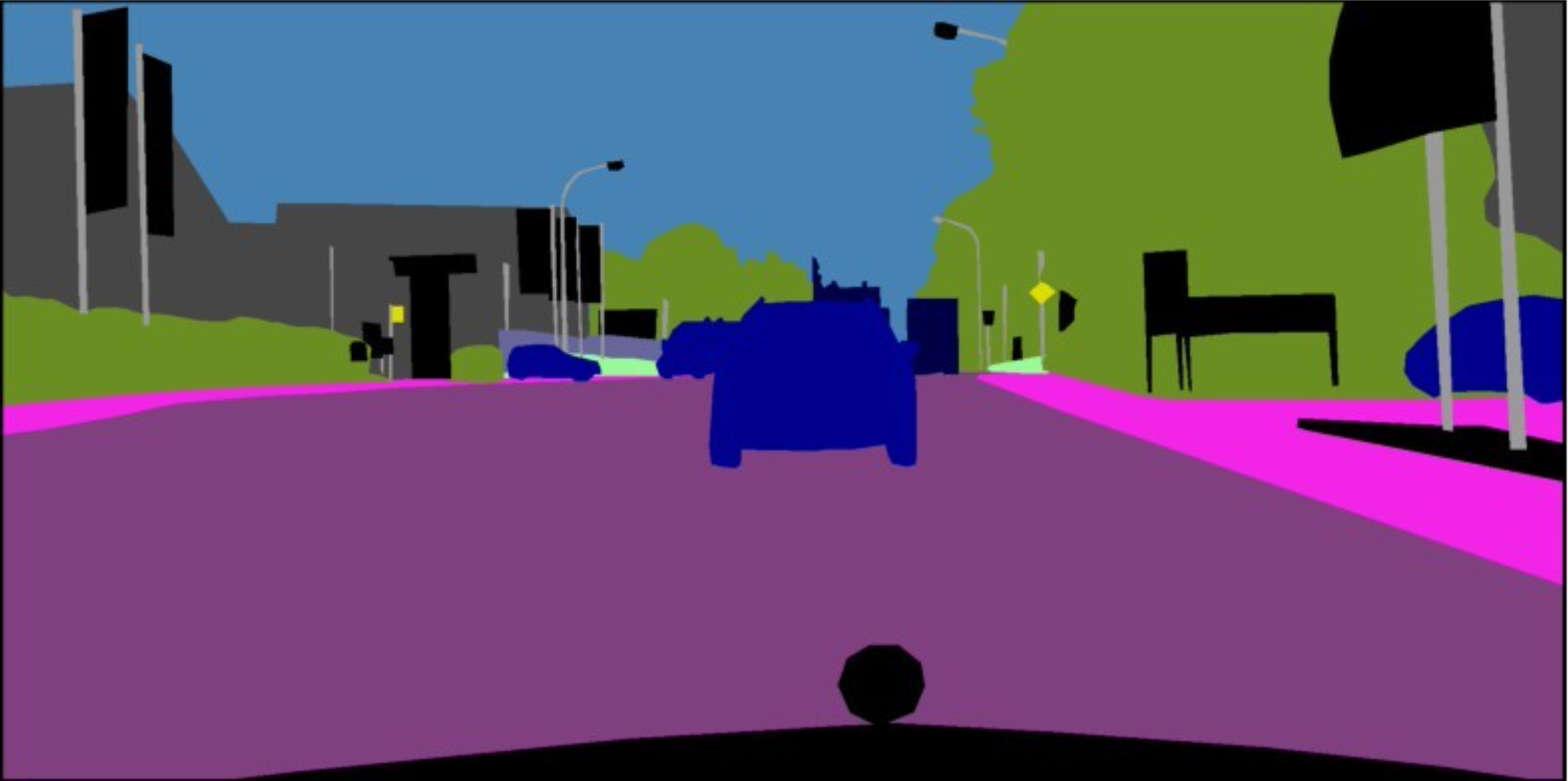}}}
\newcommand\animageda{\adjustbox{valign=m,vspace=0.2pt}{\includegraphics[width=.2\linewidth]{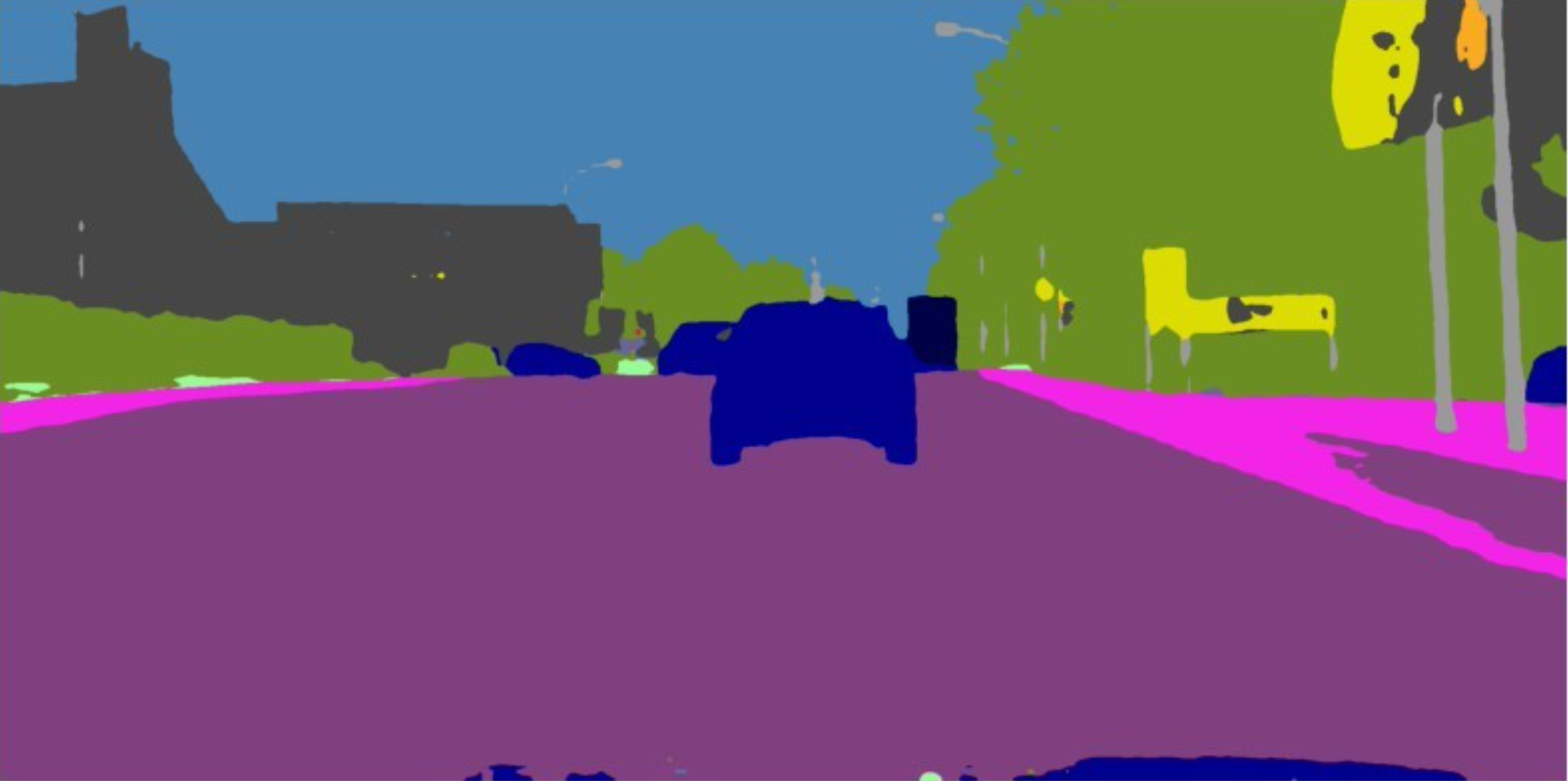}}}
\newcommand\animagehrda{\adjustbox{valign=m,vspace=0.2pt}{\includegraphics[width=.2\linewidth]{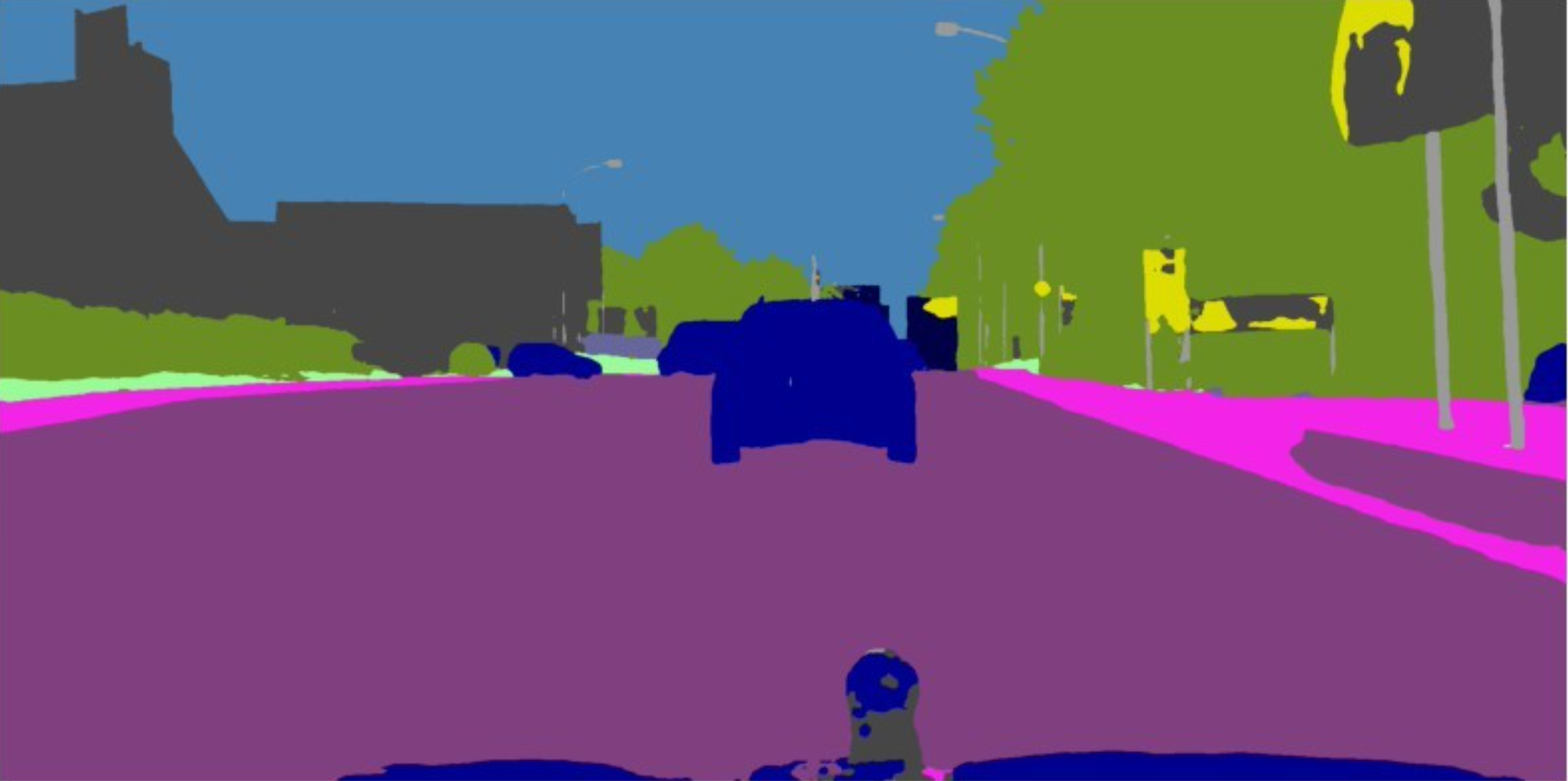}}}
\newcommand\animageSegDA{\adjustbox{valign=m,vspace=0.2pt}{\includegraphics[width=.2\linewidth]{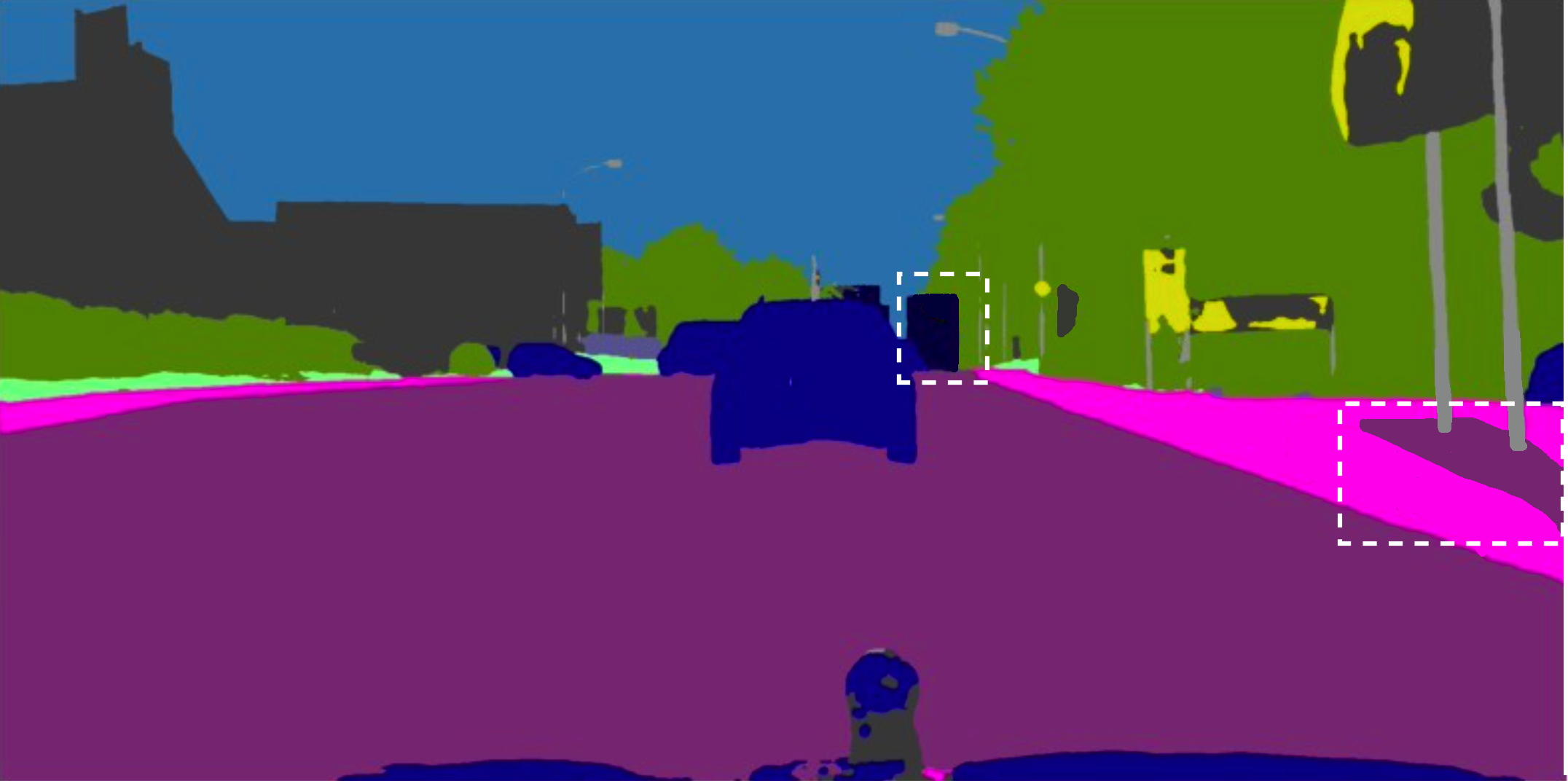}}}

\newcommand\animagetgttwo{\adjustbox{valign=m,vspace=0.2pt}{\includegraphics[width=.2\linewidth]{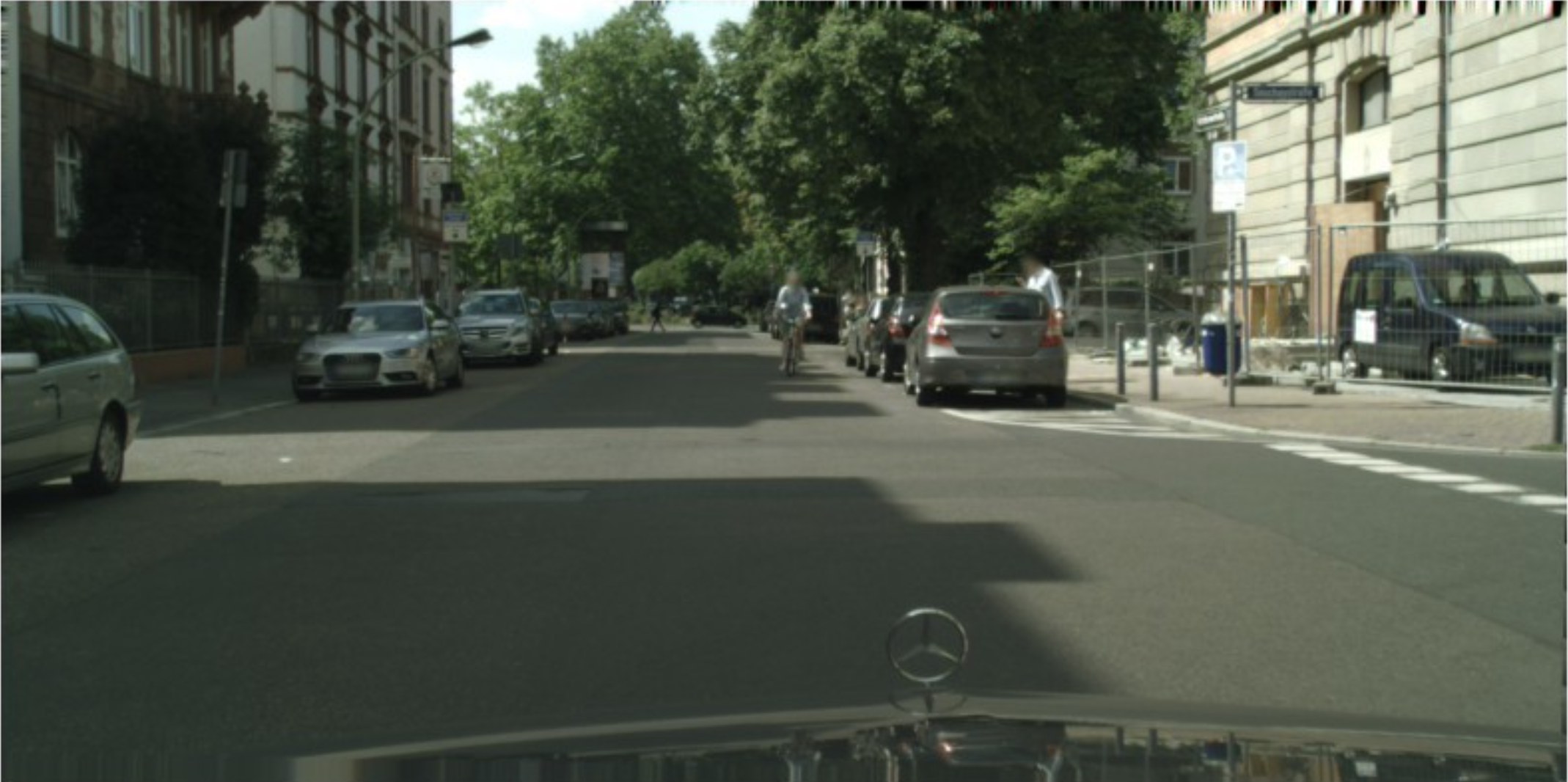}}}
\newcommand\animagegttwo{\adjustbox{valign=m,vspace=0.2pt}{\includegraphics[width=.2\linewidth]{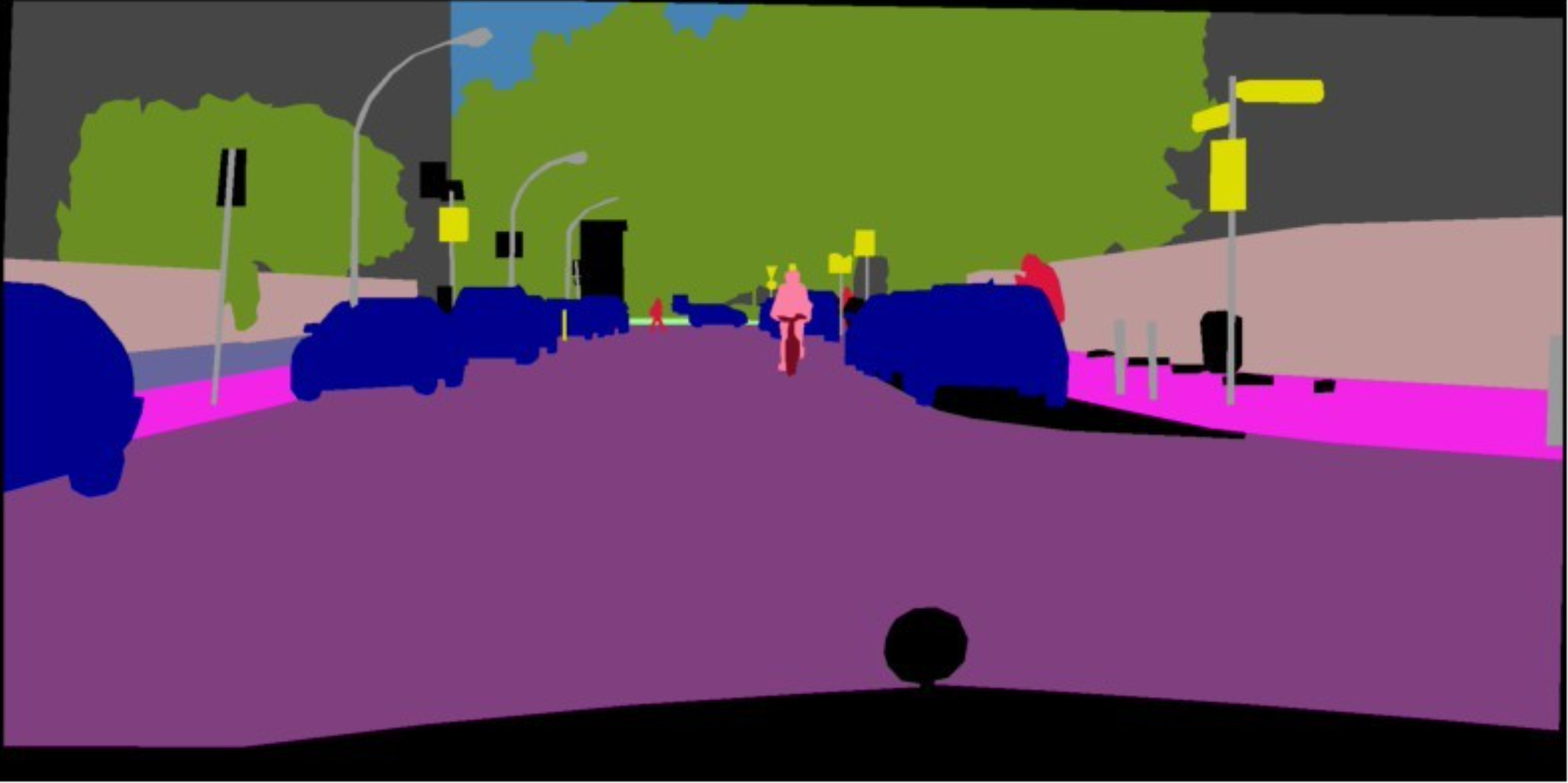}}}
\newcommand\animagedatwo{\adjustbox{valign=m,vspace=0.2pt}{\includegraphics[width=.2\linewidth]{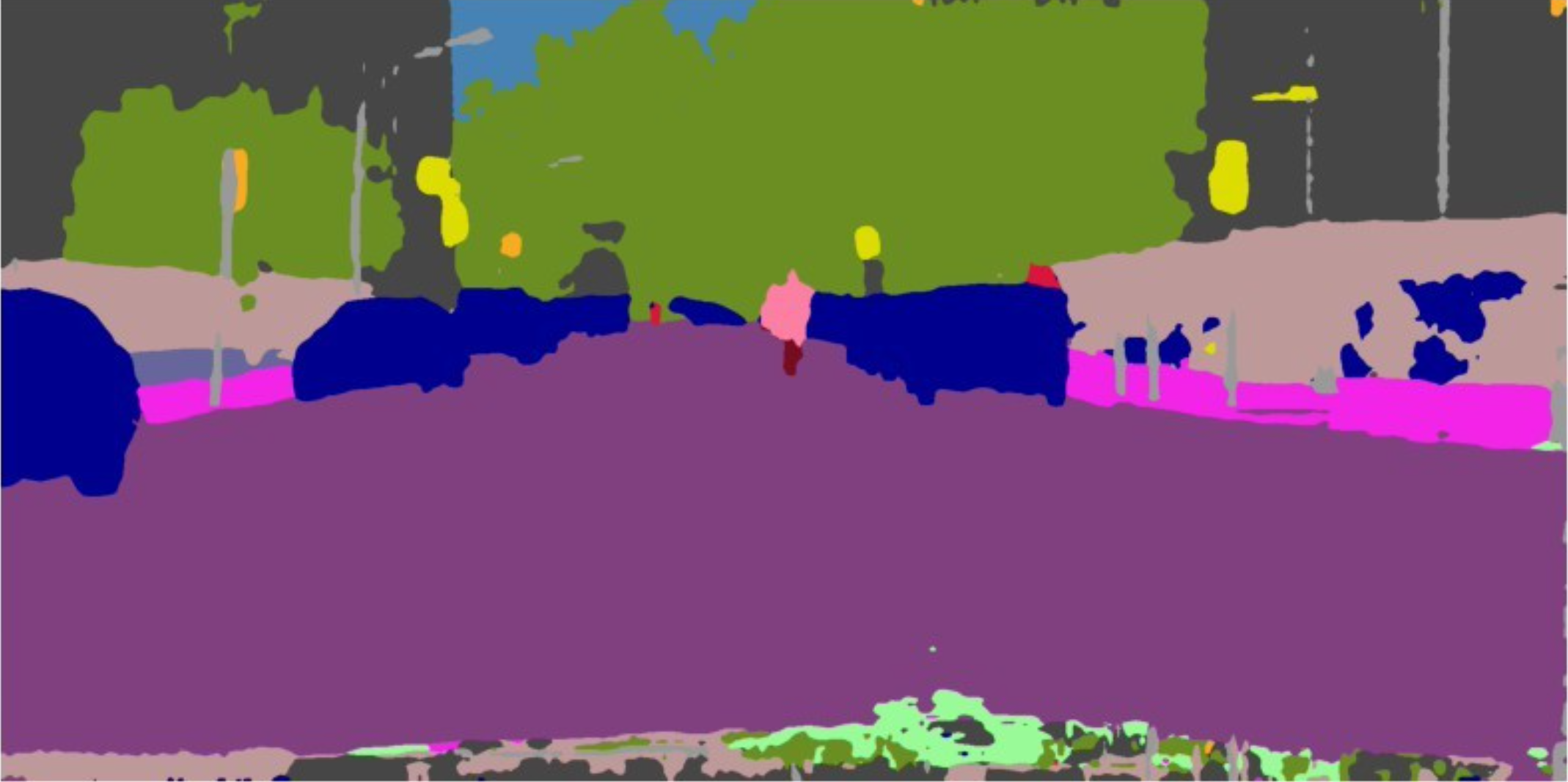}}}
\newcommand\animagehrdatwo{\adjustbox{valign=m,vspace=0.2pt}{\includegraphics[width=.2\linewidth]{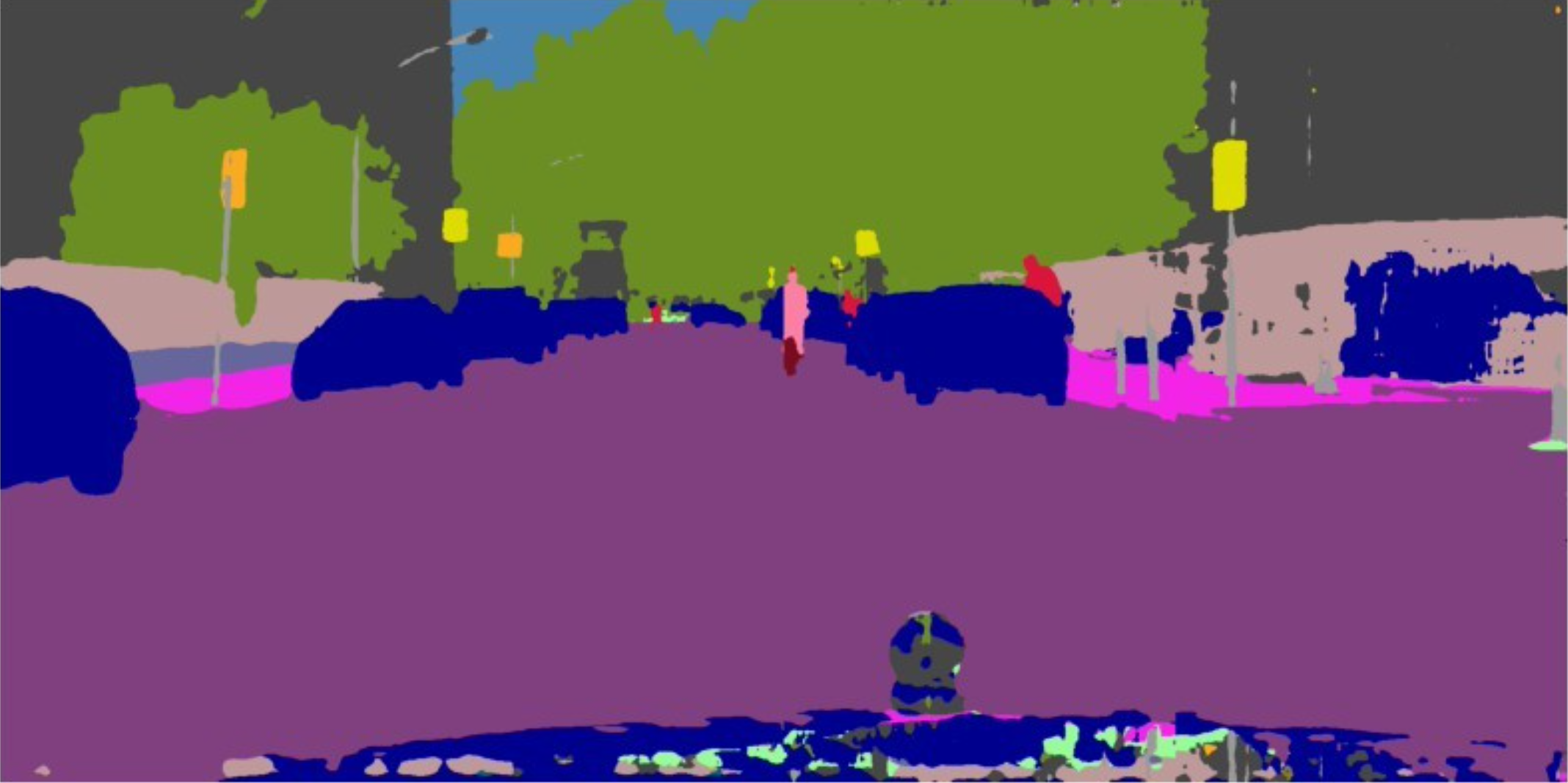}}}
\newcommand\animageSegDAtwo{\adjustbox{valign=m,vspace=0.2pt}{\includegraphics[width=.2\linewidth]{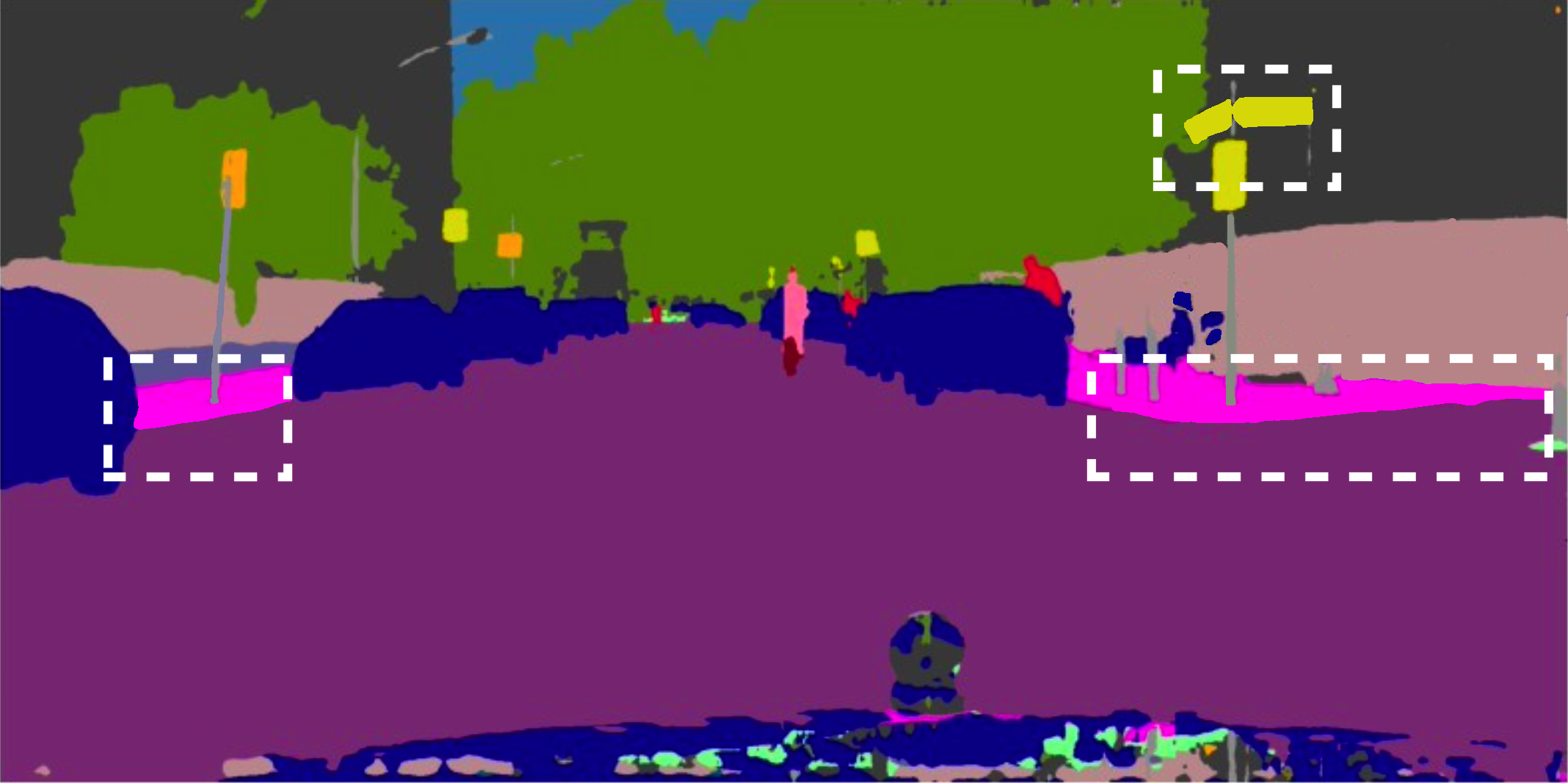}}}

\newcommand\animagetgtthree{\adjustbox{valign=m,vspace=0.2pt}{\includegraphics[width=.2\linewidth]{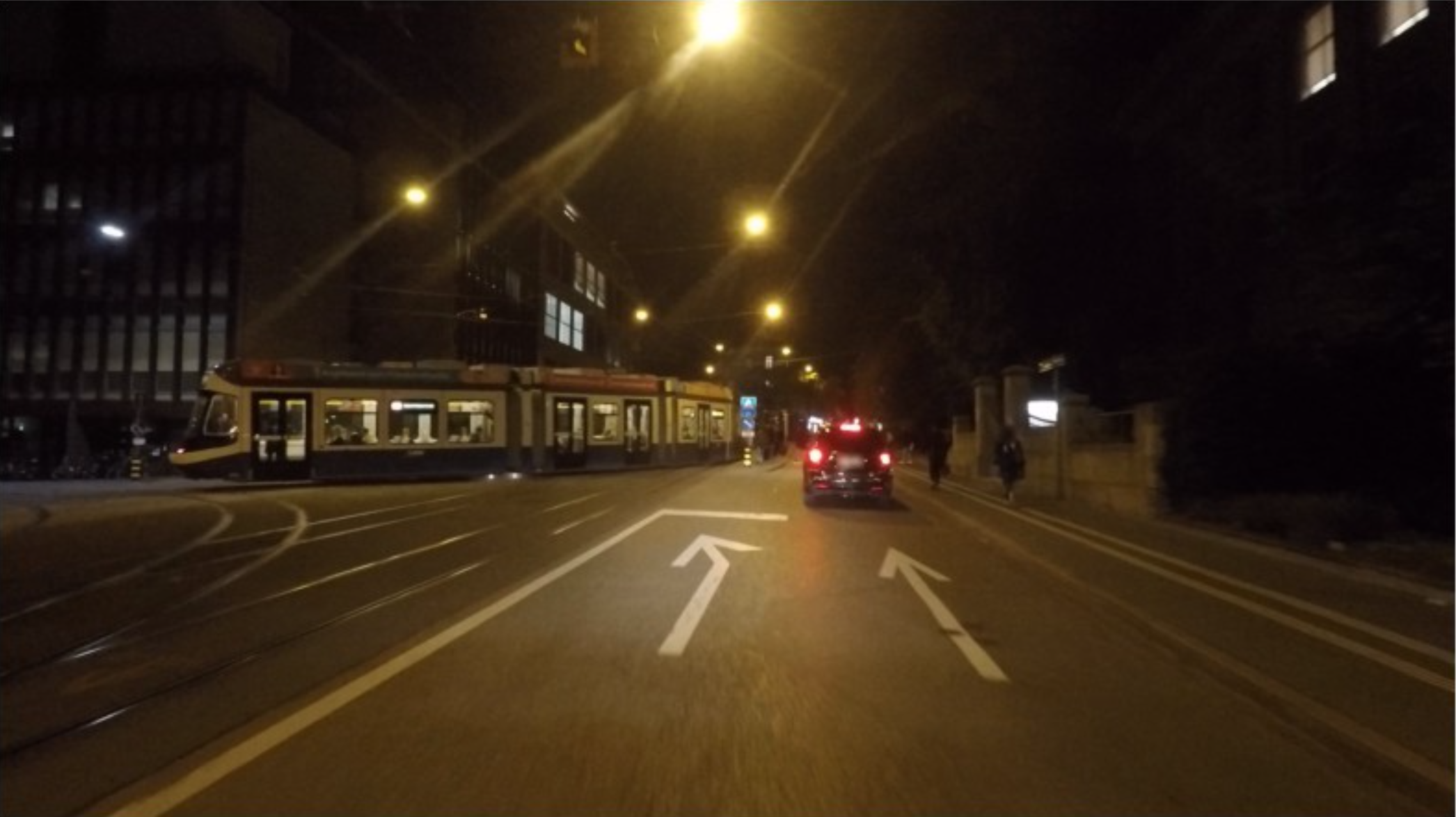}}}
\newcommand\animagegtthree{\adjustbox{valign=m,vspace=0.2pt}{\includegraphics[width=.2\linewidth]{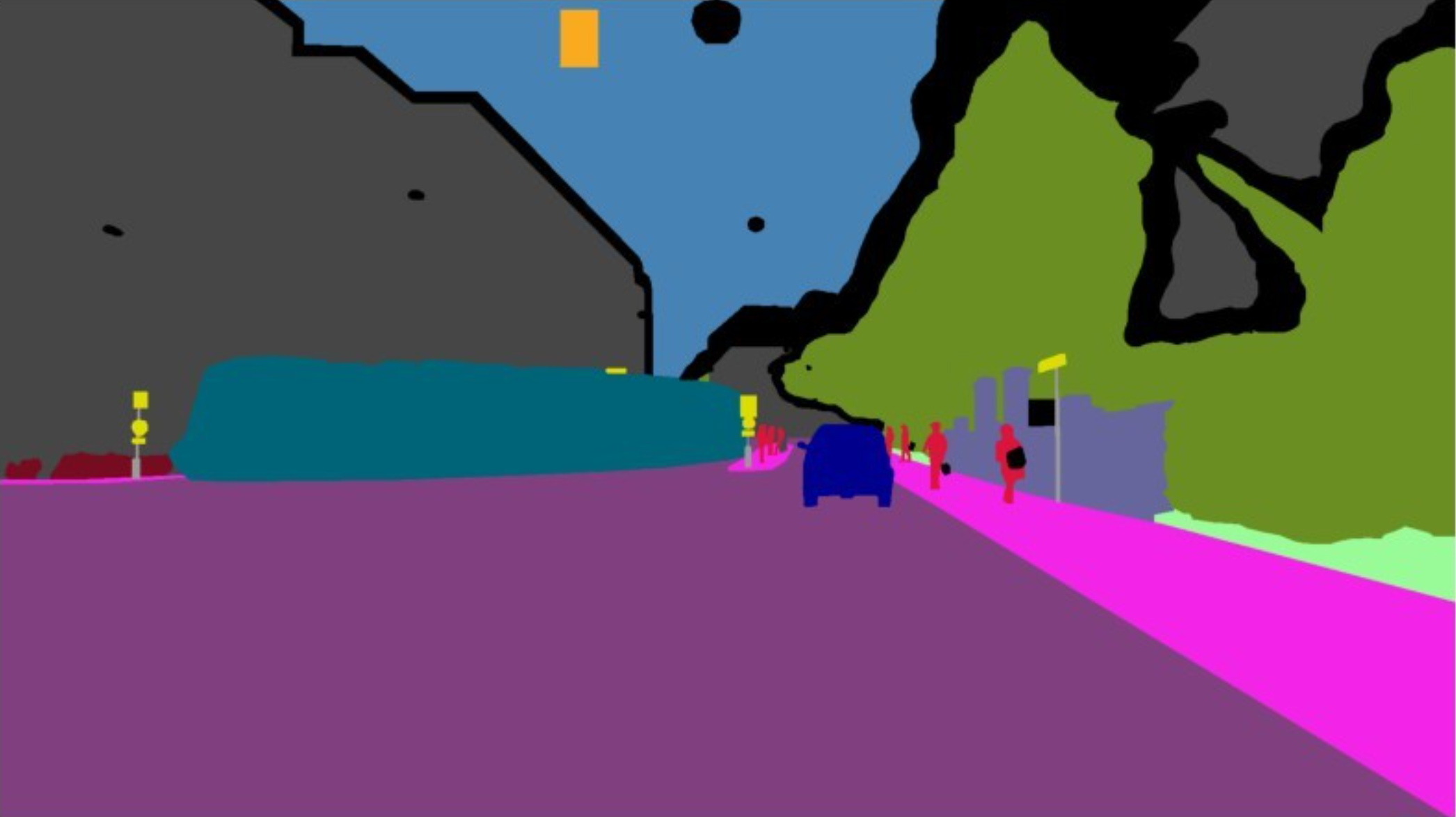}}}
\newcommand\animagedathree{\adjustbox{valign=m,vspace=0.2pt}{\includegraphics[width=.2\linewidth]{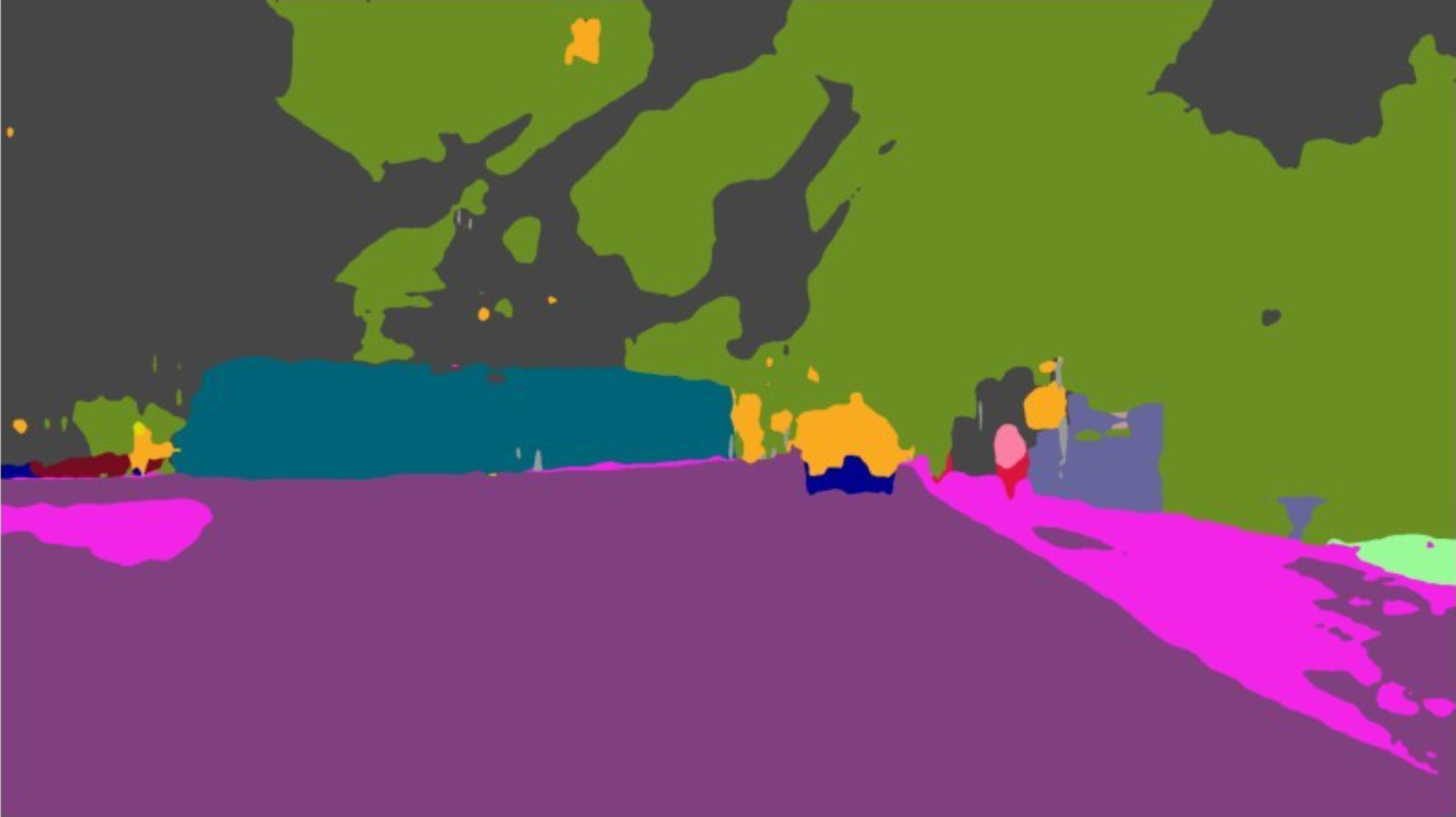}}}
\newcommand\animagehrdathree{\adjustbox{valign=m,vspace=0.2pt}{\includegraphics[width=.2\linewidth]{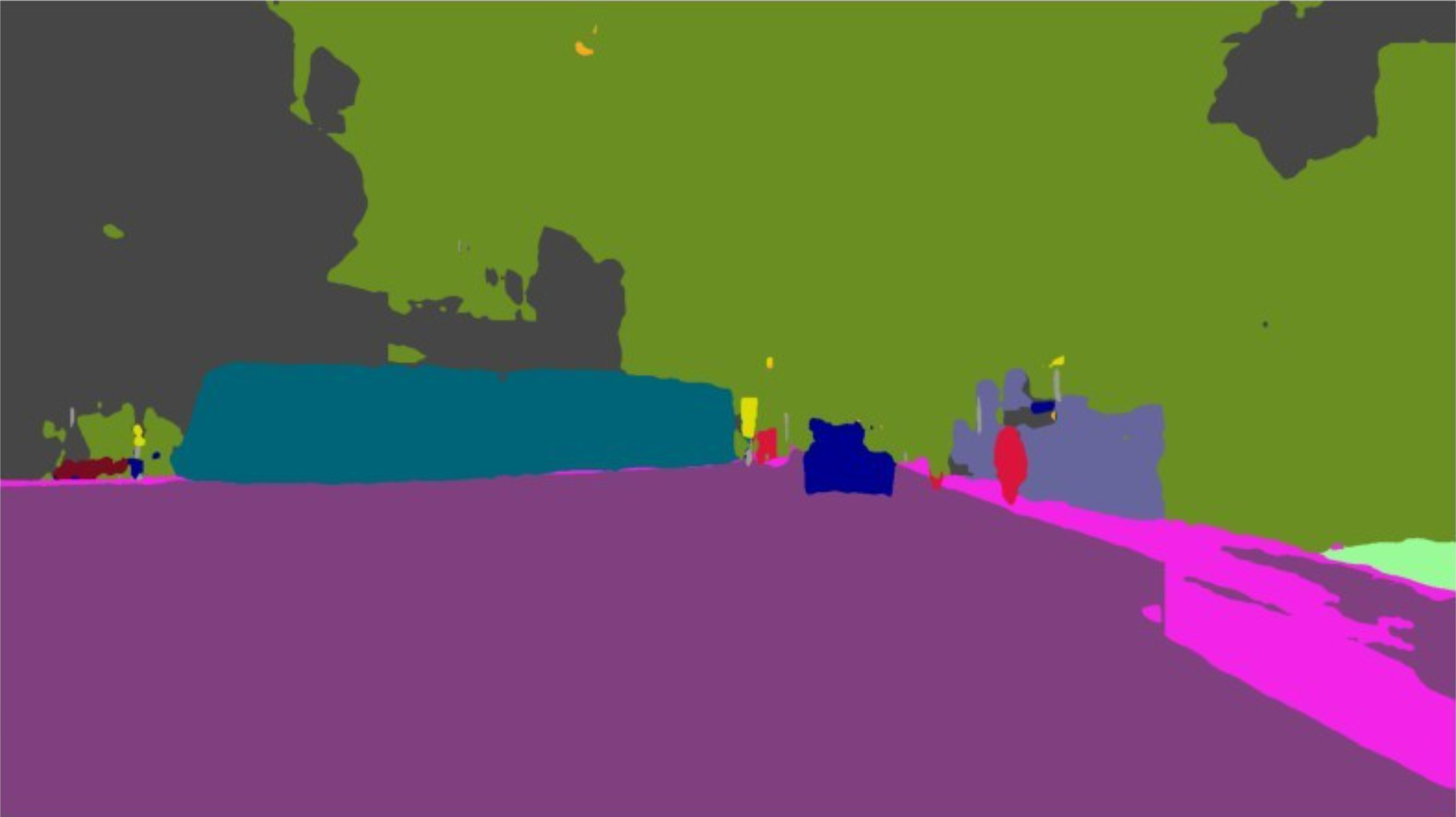}}}
\newcommand\animageSegDAthree{\adjustbox{valign=m,vspace=0.2pt}{\includegraphics[width=.2\linewidth]{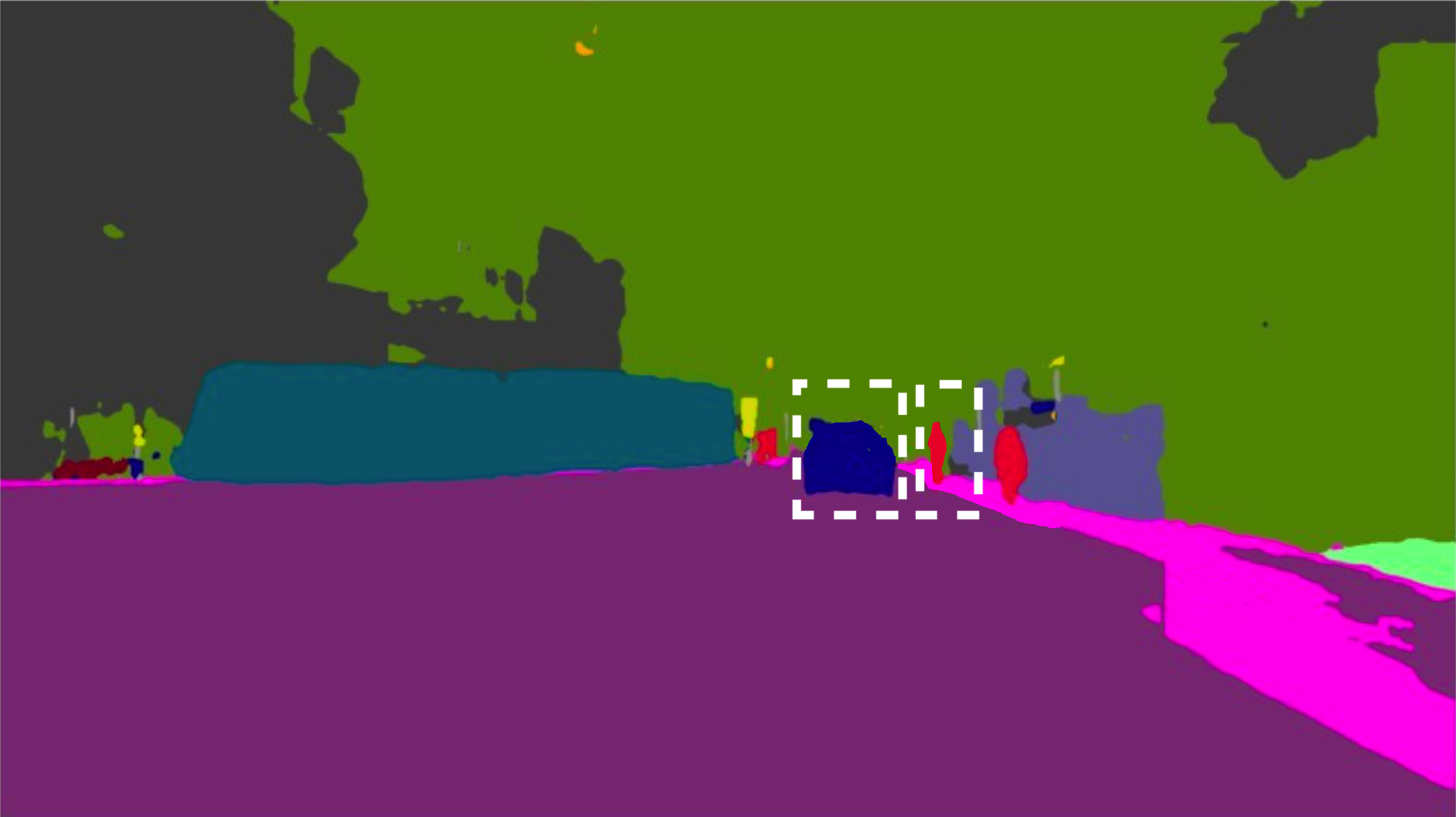}}}

\newcommand\animagetgtfour{\adjustbox{valign=m,vspace=0.2pt}{\includegraphics[width=.2\linewidth]{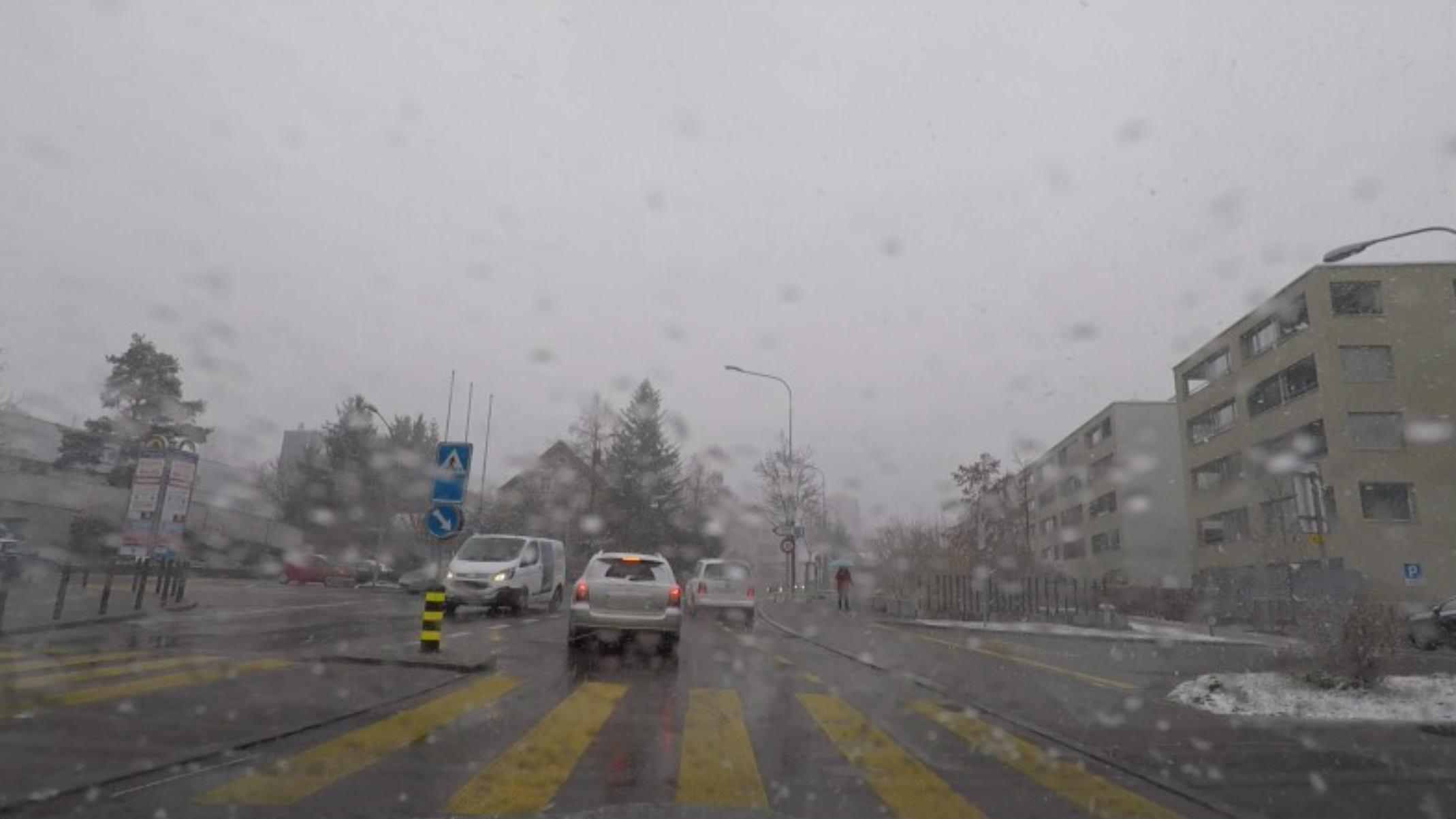}}}
\newcommand\animagegtfour{\adjustbox{valign=m,vspace=0.2pt}{\includegraphics[width=.2\linewidth]{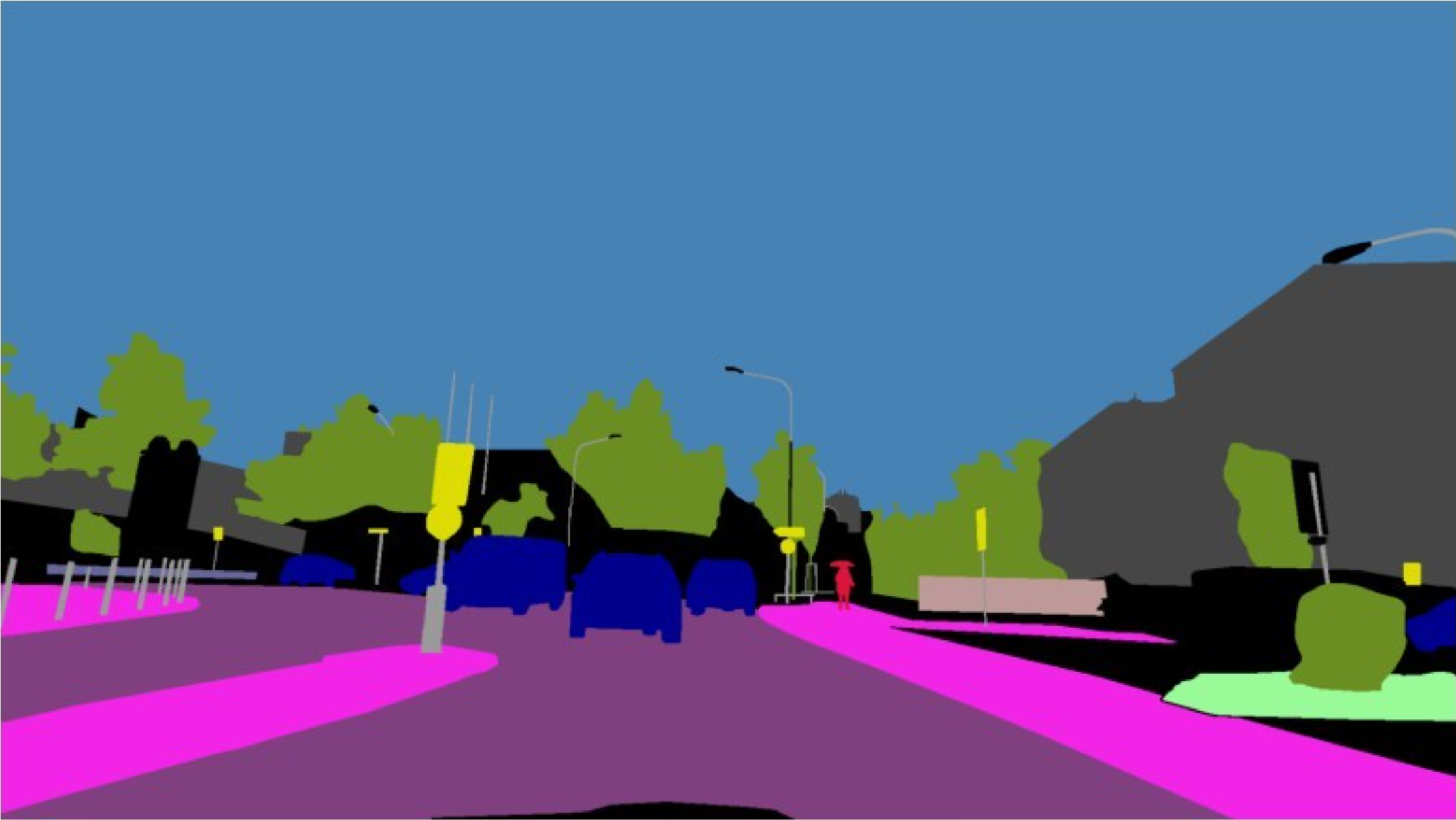}}}
\newcommand\animagedafour{\adjustbox{valign=m,vspace=0.2pt}{\includegraphics[width=.2\linewidth]{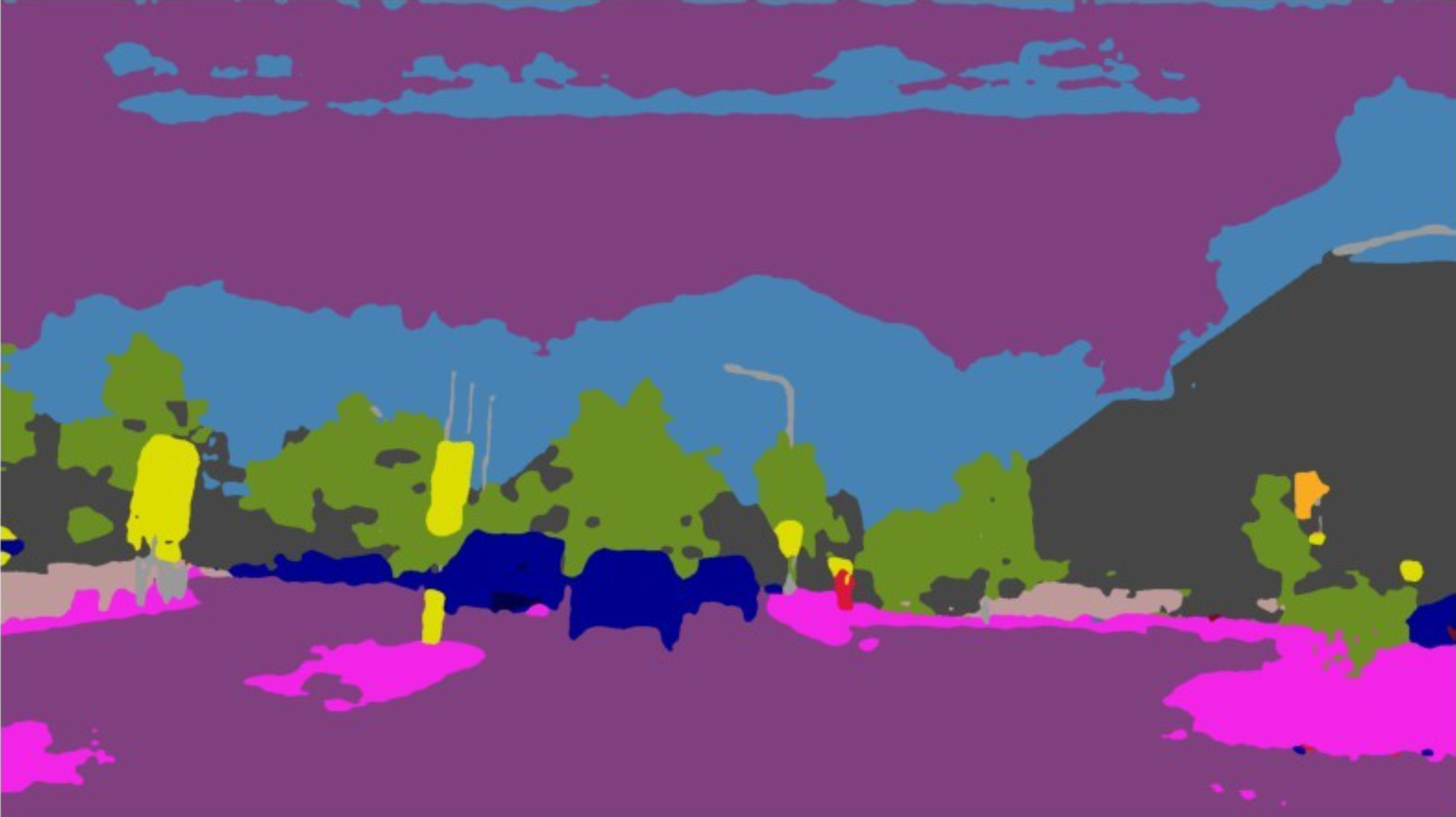}}}
\newcommand\animagehrdafour{\adjustbox{valign=m,vspace=0.2pt}{\includegraphics[width=.2\linewidth]{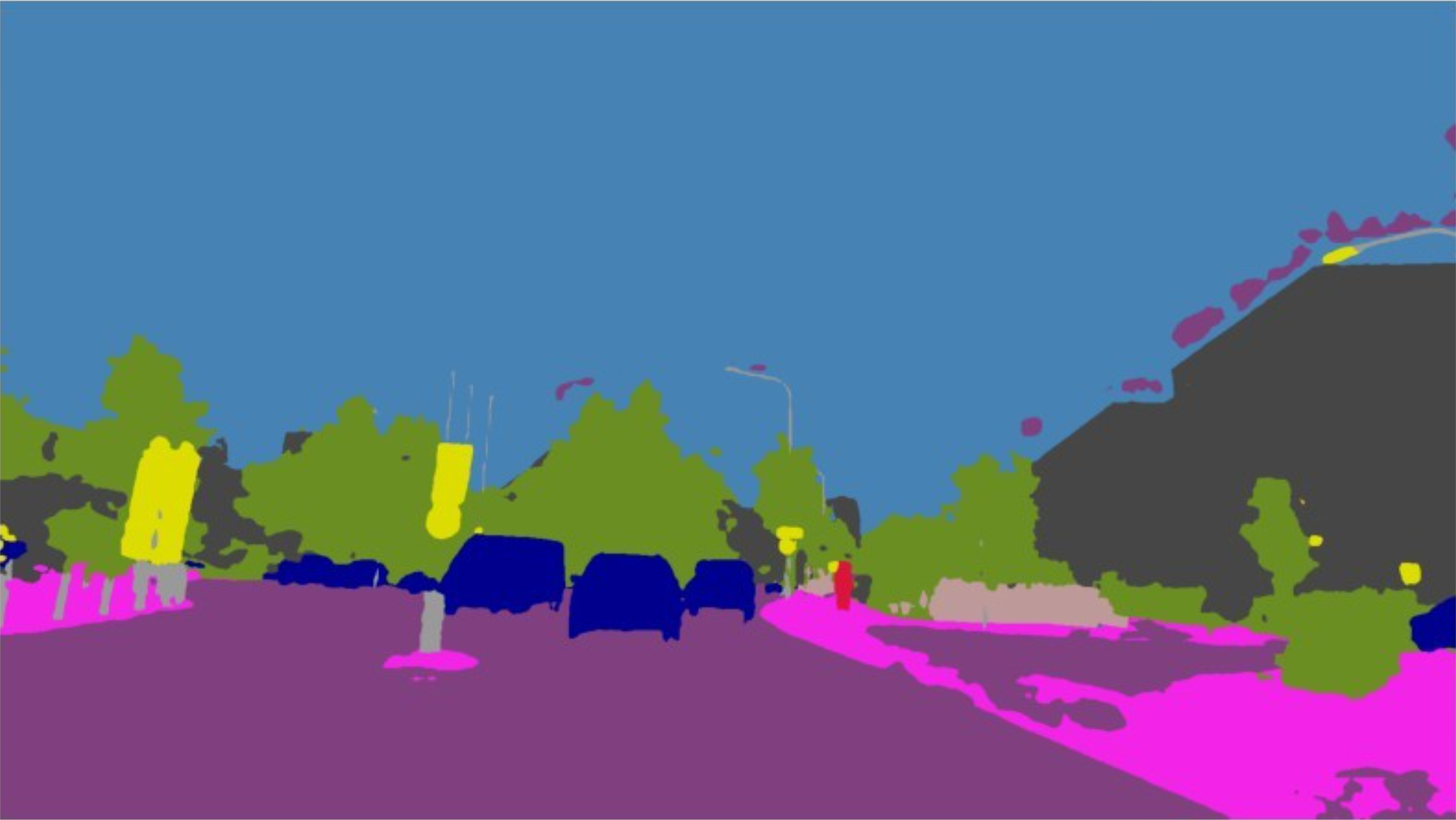}}}
\newcommand\animageSegDAfour{\adjustbox{valign=m,vspace=0.2pt}{\includegraphics[width=.2\linewidth]{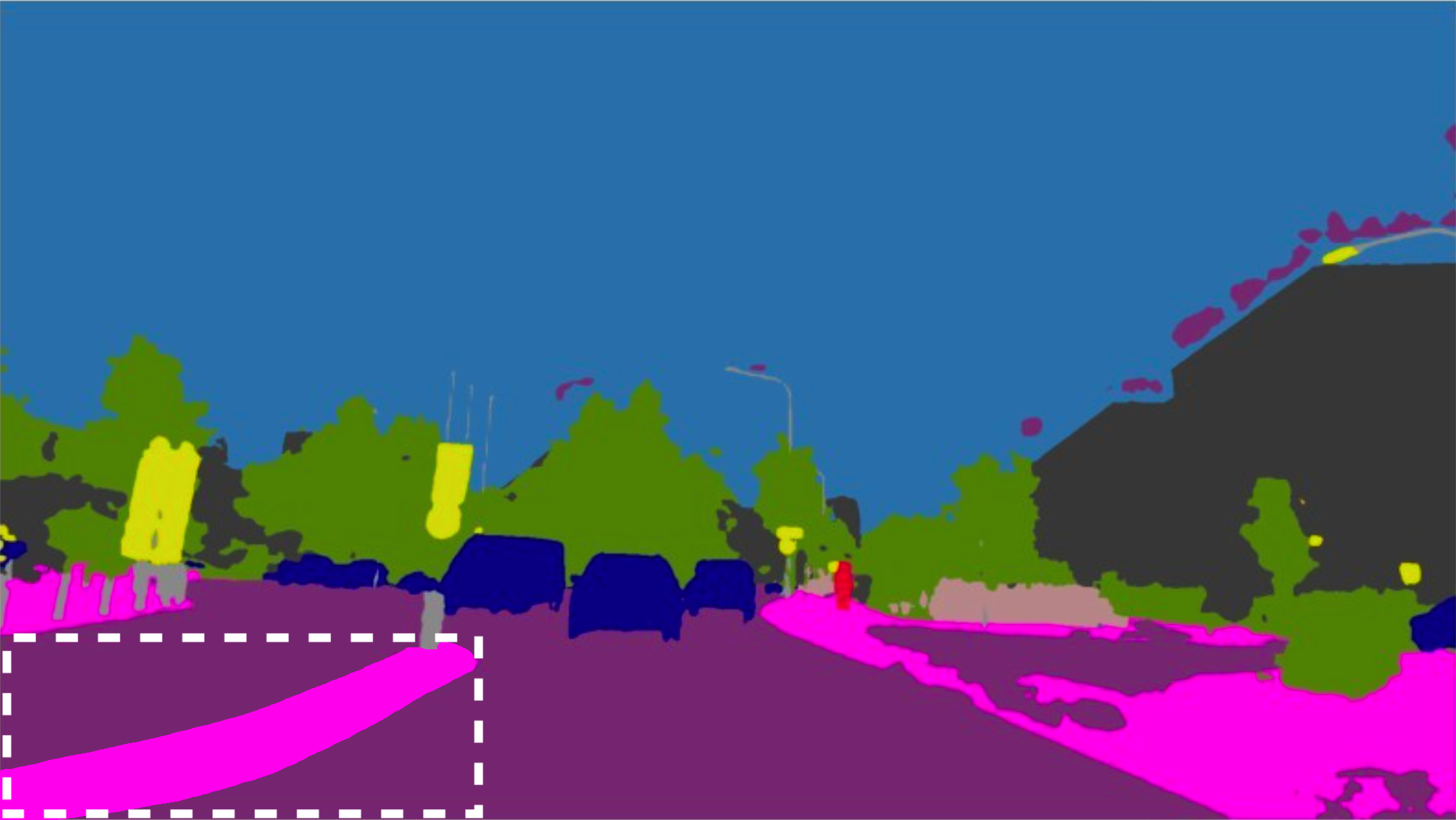}}}

\begin{document}

%%%%%%%%% TITLE
\title{SegDA: Maximum Separable Segment Mask with Pseudo Labels for Domain Adaptive Semantic Segmentation}

\author{Anant Khandelwal\thanks{Work Done while author was in Amazon}\\
Applied Scientist, Amazon\\
{\tt\small anantkha@amazon.com}
% \and
}

\maketitle
\ificcvfinal\thispagestyle{empty}\fi

%%%%%%%%% ABSTRACT
\begin{abstract}
    Unsupervised Domain Adaptation (UDA) aims to solve the problem of label scarcity of the target domain by transferring the knowledge from the label rich source domain. Usually, the source domain consists of synthetic images for which the annotation is easily obtained using the well known computer graphics techniques. However, obtaining annotation for real world images (target domain) require lot of manual annotation effort and is very time consuming because it requires per pixel annotation. To address this problem we propose SegDA module to enhance transfer performance of UDA methods by learning the maximum separable segment representation. This resolves the problem of identifying visually similar classes like pedestrian/rider, sidewalk/road etc. We leveraged Equiangular Tight Frame (ETF) classifier inspired from Neural Collapse for maximal separation between segment classes. This causes the source domain pixel representation to collapse to a single vector forming a simplex vertices which are aligned to the maximal separable ETF classifier. We use this phenomenon to propose the novel architecture for domain adaptation of segment representation for target domain. Additionally, we proposed to estimate the noise in labelling the target domain images and update the decoder for noise correction which encourages the discovery of pixels for classes not identified in pseudo labels. We have used four UDA benchmarks simulating synthetic-to-real, daytime-to-nighttime, clear-to-adverse weather scenarios. Our proposed approach outperforms +2.2 mIoU on GTA → Cityscapes, +2.0 mIoU on Synthia → Cityscapes, +5.9 mIoU on Cityscapes → DarkZurich, +2.6 mIoU on Cityscapes → ACDC.
\end{abstract}

%%%%%%%%% BODY TEXT
\section{Introduction}
With the success of Convolutional Neural Networks (CNN) \cite{chen2017deeplab, long2015fully} and Vision Transformers \cite{liu2021swin, zheng2021rethinking} based models on the task of semantic segmentation, there is an increase in interest in adopting the semantic segmentation models in production for autonomous vehicles. However, the success of these models has been shown on the synthetic datasets since obtaining per pixel annotation of synthetic datasets can be generated easily with computer graphics \cite{richter2016playing, ros2016synthia}, but obtaining these for real world is very costly since it requires lot of time for annotation of large number of images, in absence of which the deep neural network will not able to generalize for every type of scenario. There exists domain gaps between the synthetic and real world images like illumination, weather, and camera quality \cite{fang2022behavioral, wang2022multiple, wu2019ace}. To achieve the generalization on real images without any labelled dataset, researchers resort to unsupervised domain adaptation (UDA) techniques either through network changes or data augmentation to source domain (synthetic) to be able to transfer learned knowledge from source domain to target domain environment.\\
\begin{figure}
    \centering
    \includegraphics[width=\columnwidth]{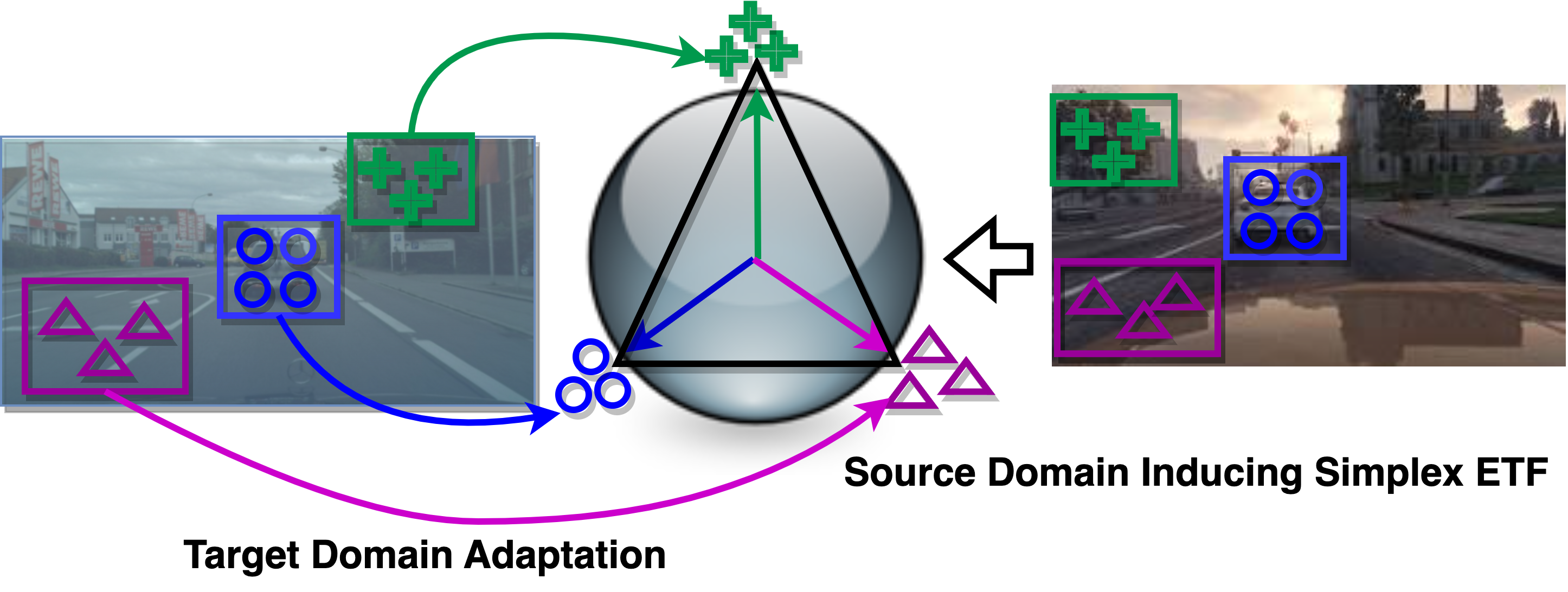}
    \caption{The idea of our proposed framework is to not only adapt the pixel representation in source domain to target domain but also make them aligned to the corresponding segment representation (collapsed representation) which is parallel to the classifier weight of corresponding classes and hence ensures maximum separability.}
    \label{fig:intro}
\end{figure}
Existing works leveraged adversarial learning \cite{luo2021category, luo2019taking, pan2020unsupervised, tsai2018learning, vu2019advent, yang2020fda}, self-supervised learning \cite{kang2020pixel, mei2020instance, tranheden2021dacs, zhang2021prototypical, zhou2022uncertainty, zou2018unsupervised, zou2019confidence} to learn domain invariant representations. Some minimizes this domain discrepancy at pixel level \cite{hoffman2018cycada, li2019bidirectional, wu2019ace, yang2020fda}, feature level \cite{huang2018domain, luo2019significance} or prediction level \cite{pan2020unsupervised, tsai2018learning, tsai2019domain, vu2019advent}. Self-Supervised learning aims to mine the visual knowledge from unlabelled images and pose the optimization objective to make these visual cues to be as closer to the ones in source domain, towards that some works have adopted augmentations to source domain like rotation \cite{komodakis2018unsupervised}, colorization \cite{zhang2016colorful}, mixup \cite{sun2022self} and random erasing \cite{zhong2020random}. However, a common issue with these methods is that their end-to-end network is very simple, relying either only on the data augmentation or techniques like use of variable dropout, teaching the network using supervision from discriminator to generate the consistent prediction between source and target domain. However, this does not resolve the problem of confusion between classes of similar visual appearances like \textit{road/sidewalk} or \textit{pedestrian/rider}. As shown in Figure \ref{fig:qual} the ground truth label of sidewalk is incorrectly predicted in the SOTA segmentation models named DAFormer\cite{hoyer2022daformer} and HRDA\cite{hoyer2022hrda}. To solve this problem we propose the use Equiangular Tight Frame (ETF classifier inspired from Neural Collapse \cite{papyan2020prevalence} ensuring the maximal separability between classes. The phenomenon states that when the neural network trained towards zero loss, the terminal layer features of each collapse to forming a ETF simplex and the corresponding collapsed feature vectors for each class is aligned with classifier weights. With the same phenomenon, we adapt the image encoder of neural network for target domain using the segment representation obtained from the target domain (along with collapsed representation of source domain) and align it with the ETF classifier weights (as shown in Figure \ref{fig:intro}). This helps to measure the noise remains in the pixel decoder and apply the noise correction training for pixel decoder. We additionally introduced the pixel discovery training for the possibility of pixel belonging to the new class and keep on introducing them via the pseudo labels obtained from the moving average based teacher network. This complete setting enables us to achieve the +2.2 mIoU on GTA $\rightarrow$ Cityscapes, +2.0 mIoU on Synthia $\rightarrow$ Cityscapes, +5.9 mIoU on Cityscapes $\rightarrow$ DarkZurich, +2.6 mIoU on Cityscapes $\rightarrow$ ACDC. The UDA benchmark of DarkZurich and ACDC correspond to images in nightitme and adverse weather conditions, achieving improvement in these UDA benchmarks proved the efficiency of our approach.

\section{Related Work}
Unsupervised Domain Adaptation (UDA) aims to solve the label scarcity problem for target domain with the successful transfer of knowledge from label rich source domain. Some CycleGAN\cite{zhu2017unpaired} based methods \cite{hoffman2018cycada, wu2018dcan} does exactly the task of visual style transfer from source domain to the target domain. These methods belong to a major category of adversarial learning methods\cite{luo2021category, luo2019taking, pan2020unsupervised, tsai2018learning, vu2019advent, yang2020fda}, which aims to learn the domain invariant representation based on min max optimization strategy, where a feature extractor is trained to fool a discriminator and thus helps to obtain the adapted feature representations. However, as shown in \cite{zheng2022adaptive}, this type of training is unstable leading to suboptimal performance. This is followed by another line works of training the segmentation network on target domain with pseudo label which can be pre-computed either offline \cite{yang2020fda, zou2018unsupervised} or updated online during training iterations \cite{hoyer2022daformer, tranheden2021dacs}. Irrespective of the way of generating pseudo labels, there is inevitable noise (due to underlying difference in data distribution between domains) in the pseudo labels which make the training noisy and leading to sub optimal performance. Some adopted the use of high confidence pseudo labels \cite{zou2019confidence, zou2018unsupervised}, some conducted domain alignment for reliable pseudo labels \cite{zheng2019unsupervised} and some works leveraging uncertainty estimation \cite{zheng2021rectifying} and efficient sampling \cite{mei2020instance}. Apart from the works mentioned above researchers also adopted combining adversarial learning and self training with specialized entropy minimization schemes\cite{chen2019domain, vu2019advent}, semantic prototype based contrastive learning method for class alignment \cite{xie2021spcl}, visual pretraining \cite{wang2021dense}, contrastive learning between features using different saliency masks \cite{van2021unsupervised}. Our proposed method is orthogonal to all the above approaches and adds value on top of SOTA methods as proved through qualitative and quantitative analysis over four UDA benchamrks. 

\section{Methods}
We start by introducing problem statement and the basic understanding of semantic segmentation along with the corresponding loss functions in supervised setting and domain adaptation setting. Following this we described our proposed model i.e. \textit{SegDA} and the modelling of maximum separable segments under the label noise implicit in pseudo labels. Finally, we discussed the the utility of loss functions in identifying the regions not highlighted in pseudo labels and the corrected loss for segmentation under label noise.\\ 
\textbf{Problem Statement}: Given the source domain data containing images $\mathcal{X}^S = \{x_k^{S}\}_{k=1}^{N_S}$, labelled by $\mathcal{Y}^S = \{y_k^{S}\}_{k=1}^{N_S}$ and the unlabelled target domain $\mathcal{X}^T = \{x_k^{T}\}_{k=1}^{N_T}$, where $N_S$ and $N_T$ are the number of images in source and target domain respectively. The label map of source domain $\mathcal{Y}^S$ contains $C$ categories. The setting for domain adaptive semantic segmentation requires to learn the function able to map the unlabelled images $\mathcal{X}^T$ to their semantic segmentation labels $\mathcal{Y}^T$ without the supervision of ground truth target domain labels. \\
\textbf{Semantic Segmentation}: A neural network supervised training on labelled images follows the existing works \cite{zheng2019unsupervised, zou2019confidence} and the supervised loss is formulated as:
\begin{equation}
    \mathcal{L}_{k}^{S} = \mathcal{H}(f_{\theta}(x_k^{S}), y_k^{S}) 
     \textrm{,                 }  \mathcal{L}^{S} = \frac{1}{N_S} \sum_{k=1}^{N_S} \mathcal{L}_k^{S}
\end{equation}
\begin{equation}
   \mathcal{H}(\hat{y}, y) = - \sum_{i=1}^{H}\sum_{i=1}^{W}\sum_{i=1}^{C} y_{ijc} \log \hat{y}_{ijc}
\end{equation}
However, this setting can only be applied to source domain where labelled data is accessible. For the target domain in the absence of ground-truth labels, predictions from source trained model on target domain does not show the similar performance as on the source domain because of underlying difference in the dataset distribution like source domain consists of synthetic images while the target domain contains the images from real world. This requires to adapt the model trained on source domain to the unlabelled target domain $\mathcal{X}^T$. Similar to \cite{tranheden2021dacs, zhou2022context} we also handled the problem of label scarcity with pseudo labels $\tilde{\mathcal{Y}}_{T} = \{\tilde{Y}_k^{T} \in \{0, 1\}^{H \times W \times C}\}_{k \in N_T}$ by teacher network $\bar{f}_{\theta}$ updated during training of student $f_{\theta}$ with exponential moving average of weights of student network at each training iteration\cite{tarvainen2017mean, zheng2022adaptive}. The loss function with pseudo labels is given as follows:
\begin{equation}
    \mathcal{L}_{k}^{T} = \mathcal{H}(f_{\theta}(x_k^{T}), \bar{Y}_k^{T}), \textrm{    } \mathcal{L}_{T} = \frac{1}{N_T} \sum_{k=1}^{N_T} \mathcal{L}_k^{T}
\end{equation}

\begin{figure*}[t]
    \centering
    \includegraphics[width=\textwidth]{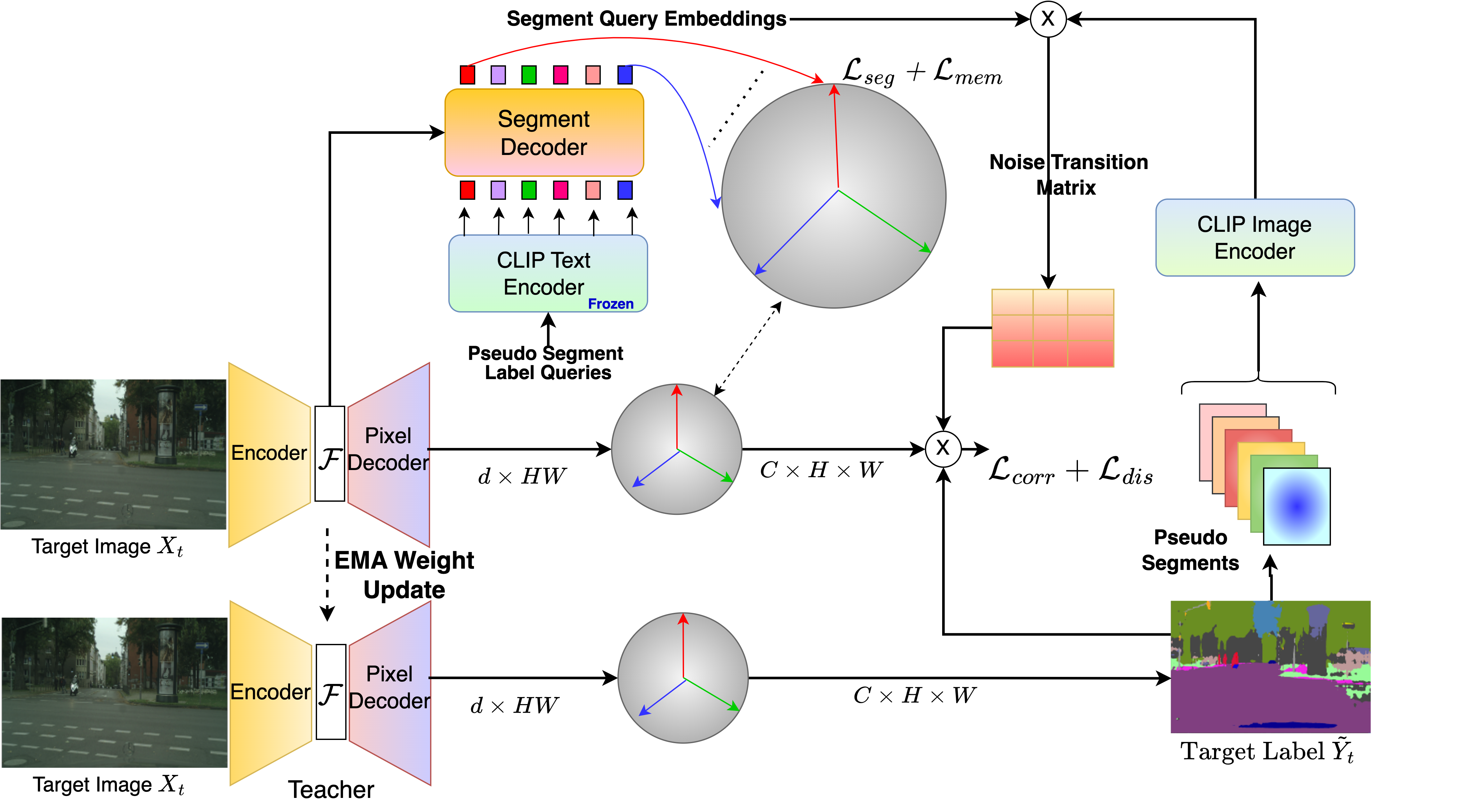}
    \caption{UDA with proposed method SegDA. The source domain trained model is adapted to target domain using domain adaptation loss $\mathcal{L}_{dapt}$, memory loss $\mathcal{L}_{mem}$ to retain the source information, noise correction loss $\mathcal{L}_{corr}$ and pixel discovery loss $\mathcal{L}_{dis}$. Overall SegDA enforces the noise correction in the pixel predictions made by an exponential moving average (EMA) teacher, where the noise is estimated using the adapted representations for each segment class which corresponds to the vector in a ETF Simplex, hence ensuring maximum separability. }
    \label{fig:SegDA}
\end{figure*}
But however the training with  pseudo labels is noisy, and hence in practice \cite{guo2022simt} we also account for only confident predictions (greater than the threshold $\tau_h$) to contribute in the loss function. However, we additionally implemented the noise estimation in the segmentation loss (denoted as $\mathcal{L}_{corr}$) for segment classes identified from pseudo labels (denoted as $C'$). To be able to discover the new segment classes not present in pseudo labels we consider the pixels having $f_{\theta}(x_k^T) < \tau_l$ (lower threshold) and $\bar{y}_k^T = \textrm{argmax} f_{\theta}(x_k^T) > C'$ incorporated in the loss function $\mathcal{L}_{dis}$. Further, we ensure the maximum separability of each of the segment using the feature collapse property of Neural Collapse \cite{papyan2020prevalence} and use the representation obtained from segment decoder to adapt the pixel classifier to the target domain representation (incorporated in the loss $\mathcal{L}_{dapt}$). To avoid forgetting the source domain, the collapsed segments representation from source domain is plugged in $\mathcal{L}_{mem}$ along with domain adaptation loss $\mathcal{L}_{dapt}$. The end-to-end network for domain adaptive semantic segmentation consists of pixel level module, segment level module and joint classifier for both pixel and segment as shown in Figure \ref{fig:SegDA}. Only pixel level module and classifier collectively called $f_{\theta}$ is the adapted semantic segmentation network and hence deployed in production for inference after domain adaptation training.\\
\textbf{Pixel-Level Module} outputs the $d$-dimensional representation for each pixel in an image of size $H \times W$. It consists of an encoder which generates the low resolution image feature map denoted as $\mathcal{F} \in R^{C_{\mathcal{E}} \times \frac{H}{S} \times \frac{W}{S}}$  where $C_{\mathcal{E}}$ is the number of channels and $S$ is the stride of the feature map. The feature map $\mathcal{F}$ is then gradually up sampled by the pixel decoder to output the $d$-dimensional pixel level feature map $\mathcal{E}_{pixel}^{d \times H \times W}$. Any existing pixel classification based segmentation model\cite{hoyer2022hrda, hoyer2022daformer, cheng2021per, chen2017deeplab, zheng2021rethinking} fits this module, but however we described the  encoder and pixel decoder outputs so as to leverage these in obtaining the segment representation (using segment decoder) and the (pixel, segment) classification using joint classifier. \\
\textbf{Segment Level Module}: To obtain the equivalent segment representation we convert the Transformer decoder\cite{vaswani2017attention} with N- positional embeddings as queries to the decoder with text embeddings obtained from CLIP text encoder \cite{radford2021learning} as queries. Specifically, the text embeddings are calculated for $C'$ segment labels identified from the pseudo labels. These $C'$ embeddings at each position as query is not trainable and the image features $\mathcal{F}$ as key and values are used to generate the segment representation $\mathcal{S}^{d_{S} \times C'}$ at the output of transformer decoder.\\ 
\textbf{Joint Pixel and Segment Classifier}: We proposed the neural collapse \cite{papyan2020prevalence} inspired classifier capable of classifying both pixels and segment to the segment classes. Recent works have studied the practice of training DNN towards zero loss, this reveals that the classifier weights and last layer features collapse to form a geometric structure in the form of Equiangular Tight Frame (ETF). Essentially, the properties is stated as follows:
\begin{itemize}
    \item ($\mathcal{NC}1$) \textbf{Variability Collapse}: Last layer features of a class collapse into within-class mean. 
    \item ($\mathcal{NC}2$) \textbf{Convergence}: The within class means of all the classes converge to a vertices of a simplex ETF. 
    \item ($\mathcal{NC}3$) \textbf{Classifier Convergence}: Within-class means aligned to their corresponding classifier weights and hence classifier will also converge to form a simplex ETF. 
\end{itemize}
Neural collapse describes the optimal geometric structure of the classifier, following \cite{papyan2020prevalence} we pre-fixed this optimality by fixing the learnable classifier structure to the simplex ETF. Therefore the segmentation network $f_{\theta}$ is consists of the pixel-level module denoted by $f_{\theta}^P \in R^{d \times H \times W}$ and the classifier $W_{ETF} \in R^{d \times C}$. The classifier weights are then initialized as per the simplex representation given as:
\begin{equation}
    W_{ETF} = \sqrt{\frac{C}{C-1}} \mathbf{U}(\mathbf{I}_K - \frac{1}{C} \mathbf{1}_C \mathbf{1}_C^T)
\end{equation}
where $W_{ETF} = [\mathbf{w}_1, \mathbf{w}_2, \mathbf{w}_3.......w_C] \in R^{d \times C}$, $\mathbf{U} \in R^{d \times C}$ allows the rotation, and satisfies $\mathbf{U}^T\mathbf{U} = \mathbf{I}_C$, $\mathbf{I}_C$ is an identity matrix and $\mathbf{1}_C$ is an all ones vector. This initialization offers $W_{ETF}$ to be maximally pairwise separable. For any pair $({c}_1, {c}_{2})$ of classifier $W_{ETF}$ satisfies:
\begin{equation}
    \mathbf{w}_{{c}_1}^T \mathbf{w}_{{c}_{2}} = \frac{C}{C -1} \delta_{c_1, c_2} - \frac{1}{C - 1}, \textrm{    } \forall (c_1, c_2) \in [1, C]
\end{equation}
During domain adaptation the source and target classes will remain same and hence the classifier prototypes trained for source domain and adapted for target domain. We utilize the dot-regression (DR) loss \cite{yang2022we} for source domain training with $W_{ETF}$ instead of cross-entropy (CE) loss since CE contains both the \textsc{push} and \textsc{pull} term, where \textsc{push} term separates the feature vector of a class with classifier prototypes of different classes (but is inaccurate as highlighted in \cite{papyan2020prevalence}) there we live only with \textsc{pull} term which bring closer the feature vector of a class and the corresponding classifier prototype. The DR loss is formulated as:
\begin{equation}
    \underset{\theta_P}{\textrm{min }}\mathcal{L}(f_{{\theta}_{i}}^P, W_{ETF}) = \frac{1}{2}(\mathbf{w}_{c_i}^Tf_{{\theta}_{i}}^P - 1)^2
\label{dr_loss}
\end{equation}
where $\mathbf{w}_{c_i}^T$, is the classifier prototype corresponding to class $c_i$ and $\theta_P$ are the parameters of pixel module. The feature vector for each pixel is batch normalized using the batch normalization layer as the last layer. The loss in equation \ref{dr_loss} is summed over each batch input $x_i$. The gradient of loss in equation \ref{dr_loss} 
 ($f_{{\theta}_{i}}^P$ is considered as optimization variables as in \cite{dang2023neural}) ${\partial \mathcal{L}}/{\partial f_{{\theta}_{i}}^P} = - (1 - \cos \angle (f_{{\theta}_{i}}^P, \mathbf{w}_{c_i}))\mathbf{w}_{c_i}$, which is effectively pulling the feature towards the classifier prototype for class $c_i$ and hence converge to the simplex ETF classifier weights $W_{ETF}$ resulting in collapsed representation for each class. The prediction score for all classes for a particular pixel representation is given as $\left \langle f_{{\theta}_{i}}^P, \mathbf{w}_{c_k}\right \rangle \forall c_k \in [1, C]$, results in predicted feature map $\mathcal{E}_{pred}^{C \times H \times W}$. 
 To be able to domain adapt the classifier without forgetting the source domain we maintain the memory for each class collapsed features as the mean of all representations as per variability collapse in $\mathcal{NC}1$. 
 \begin{equation}
     \mathcal{M}_{c_i} = \frac{1}{N_{c_i}}\sum_{i=1}^{N_{c_i}} f_{{\theta}_{i}}^P
     \label{mem}
 \end{equation}
$N_{c_i}$ denotes the number of samples (pixels) across all the images in the training dataset. This results in memory $\mathcal{M} = [\mathcal{M}_{c_1}, \mathcal{M}_{c_2}, ......., \mathcal{M}_{C}]$. These memory vectors along the target domain segment representations are then used to adapt the classifier prototypes for each segment without forgetting the source domain.
\subsection{Segment Representation Adaptation}
The feature for every pixel belonging to the corresponding segment class collapse to within-class means, this collapsed representation is effectively the segment representation corresponding to each segment class, and hence the memory representations we have obtained for the source domain in equation \ref{mem} are the segment representation for corresponding class in the source domain. These source representations are aligned with the corresponding classifier prototypes of $W_{ETF}$. We adapt the target domain segment representations (obtained from segment level module) using these classifier prototypes. The idea of introducing the segment module is to denoise the pixel decoder to obtain pixel representation for target domain without forgetting the source domain and eliminating the requirement of keeping all source domain samples in the memory. For each of the pseudo label class $c \in [1, C']$ (identified in the input image from target domain) their corresponding label representation from CLIP Text Encoder \cite{radford2021learning} is obtained and used as query in the segment decoder as shown in Fig. \ref{fig:SegDA}. The domain adaptation loss for segment representation is formulated as:
\begin{equation}
    \underset{\theta_{e}, \theta_{sd}}{\textrm{min }}\mathcal{L}_{dapt} = \frac{1}{2}(\mathbf{w}_{c_i}^T\mathcal{S}_{c_i} - 1)^2, \textrm{  } \forall c_i \in [1, C']
    \label{dapt}
\end{equation}
where $\theta_{e}, \theta_{sd}$ are the parameters for encoder and segment decoder respectively. This loss will be summed over each batch containing sample $x_i$ from target domain images. Since the training of encoder and segment decoder with target domain completely wipes out the source domain information, we add the memory loss formulated as:
\begin{equation}
    \underset{\theta_P}{\textrm{min }}\mathcal{L}_{mem} = \frac{1}{2}(\mathbf{w}_{c_i}^T \mathcal{M}_{c_i} - 1)^2, \textrm{  } \forall c_i \in [1, C']
\label{mem}
\end{equation}
where parameters $\theta_P$ comprises the encoder parameters $\theta_e$ and pixel decoder parameters $\theta_{pd}$.  

\subsection{Noise Estimation and Pixel Class Discovery}
Training with memory loss and adaptation loss ensures the encoder to retain the information for source domain along with learning for target domain. However, pixel decoder contains only the source domain information and hence produces noisy pixel class distributions for target domain and hence it requires to denoise the pixel decoder. The error can eb of two types: 1) the noise between the classes $c \in [1, C']$ identified from the pseudo labels and 2) the noise due to incorrect prediction where the actual ground truth belongs to class category outside $C'$. For (1) we estimate the noise transition matrix for each segment class $c \in [1, C']$ identified from the pseudo labels. For (2) we propose a new loss which facilitates the discovery of pixels belong to class category $c \in [C'+1, C]$.  For each of segment masks identified from the pseudo labels we cropped the $C'$ images from the predicted segments in each image. We obtain the noisy segment representation $\mathcal{S}_{noisy}^{d \times C'}$ from the cropped images for target domain using CLIP Image Encoder \cite{radford2021learning}, where $d$ is the embedding dimension. Effectively the noise transition matrix is given as $\mathcal{N} = \mathcal{S}^T\mathcal{S}_{noisy} \in R^{C' \times C'}$, where $d_S = d$. For the pseudo labels obtained from the mean teacher we use the max across class scores to predict the best class if it is greater than high threshold $\tau_h$ given as, $\textrm{argmax}_{c_k} \left \langle  e_{c_k} > \tau_h \right \rangle \forall e_{c_k} \in \mathcal{E}_{pred}^{C \times H \times W}$ effectively resulting in $C'$ classes forming different segments in a given image $\mathbf{x}_k$ denoted by $\tilde{y}_k^{T} \in \{0,1\}^{H \times W \times C'}$. For these $C'$ classes the student predictions is denoted as $\mathcal{E}_{st, C'}$. Only the predictions belong to $C'$ classes (from the student) can be noise corrected hence $\mathcal{E}_{st, C'}$ is noise corrected as ($\mathcal{N} \cdot \mathcal{E}_{st, C'}$), and the corrected loss is formulated as:
\begin{equation}
   \mathcal{L}_{corr} = - \sum_{k \in N_T} 
   \mathcal{H}(\mathcal{N} \cdot \mathcal{E}_{st, C'}, \tilde{y}_k^{T}) \textrm{ } \forall c \in [1, C']
   %\tilde{Y}_k^{T}\log(\mathcal{N} \cdot \mathcal{E}_{st, C'}) 
\label{corr}
\end{equation}
We want to ensure the discovery of pixels belonging to the classes outside $C'$ and hence we consider the pixel scores from teacher $e_{c_k} < \tau_l$ (lower threshold) and the one-hot pseudo label is $\tilde{y}_k^o = \textrm{argmax}_{c_k} (e_{c_k}) > C'$. The threshold $\tau_h = 0.8 \textrm{ and } \tau_l = 0.2$ in our case. The pixel discovery loss is then formulated as:
\begin{equation}
    \mathcal{L}_{dis} = -\sum_{k \in N_T} \mathcal{H}(\mathcal{E}_{st} \setminus \tilde{y}_k^{T}, \tilde{y}_k^o) \textrm
{ } \forall c \in [C' + 1, C]
    %\tilde{y}_k^o \log(\mathcal{E}_{st} \setminus \tilde{Y}_k^{T})
    \label{dis}
\end{equation}
where $\mathcal{E}_{st} \setminus \tilde{Y}_k^{T}$ denotes the student predictions outside the $C'$ classes.
\subsection{Overall Optimization Scheme}
After training the pixel level module for source domain governed by loss given in equation \ref{dr_loss}, the source trained pixel module is then adapted to target domain governed by loss functions given in equation \ref{dapt}, \ref{mem}, \ref{corr} and \ref{dis}. The combined loss is given as:
\begin{equation}
    \underset{\theta_P, \theta_{sd}}{\textrm{min }} \mathcal{L}_{dapt} + \mathcal{L}_{mem} + \mathcal{L}_{corr} + \mathcal{L}_{dis}
\end{equation}
The teacher network $\phi_P$ (pixel module) is implemented as an EMA teacher \cite{tarvainen2017mean}. Its weights are the exponential moving average (EMA) of the weights of the (student) network $\theta_P$. 
\begin{equation}
    \phi_{P, t + 1} \leftarrow \alpha \phi_{P, t}  + (1 - \alpha) \theta_{P,t}
\end{equation}
where $t$ is the training step. The EMA teacher effectively an ensemble of student models at different training steps, which is a most widely used learning strategy in semi-supervised setting \cite{french2019semi, hoyer2021three, sohn2020fixmatch, tarvainen2017mean} and UDA \cite{araslanov2021self, hoyer2022daformer, hoyer2022hrda, tranheden2021dacs}. As the training grows the teacher is updated from student $\theta_P$ obtaining more context of what could be the stable pseudo labels based on the noise correction loss and the segment adaptation resulting in increased domain adapted performance on target domain. 
\section{Experiments}
\textbf{Datasets}: We study the domain adaptation setting considering various realistic scenarios for street scenes i.e. synthetic-to-real, clear-to-adverse weather, and day-to-nighttime. There are public datasets available for both synthetic as well as realistic environments. For synthetic dataset, we use GTA \cite{richter2016playing} containing 24,966 training images ($1914 \times 1052$ pixels) and Synthia \cite{ros2016synthia} containing 9,400 images ($1280 \times 760$ pixels). For clear weather, we use  Cityscapes (CS) \cite{cordts2016cityscapes} consisting of 2,975 and 500 images ($2048 \times 1024$ pixels) for training and validation respectively. For nighttime we use DarkZurich \cite{sakaridis2020map} with 2,416 and 151 images ($1920 × 1080$ pixels) for training and test respectively. For adverse weather (fog, night, rain, and snow) we use ACDC \cite{sakaridis2021acdc} containing 1,600, 406 and 2,000 images ($1920 \times 1080$ pixels) for training, validation and test respectively. The training resolution as per the used UDA pixel level module (half-resolution for DAFormer \cite{hoyer2022daformer} and full resolution for HRDA \cite{hoyer2022hrda}).\\
\textbf{Structure Details}: We adopt the pixel level module following recent SOTA UDA setting \cite{hoyer2022daformer, xie2023sepico, zhou2022context} based on DAFormer network \cite{hoyer2022daformer} consists of a MiT-B5 encoder \cite{hoyer2022daformer, xie2021segformer} pretrained on ImageNet-1k \cite{deng2009imagenet}. Following HRDA \cite{hoyer2022hrda} we used the context aware feature fusion decoder (from DAFormer embedding dimension 768) and for scale attention decoder we use SegFormer MLP decoder \cite{xie2021segformer} with an embedding dimension of 768 matching the dimension of Segment Decoder (Transformer Decoder \cite{vaswani2017attention} as in DETR\cite{carion2020end}) and CLIP Image Encoder \cite{radford2021learning}. Specifically, to compare \textbf{SegDA} on various setting we used DAFormer\cite{hoyer2022daformer}, HRDA\cite{hoyer2022hrda} and a DeepLabV2 \cite{chen2017deeplab} (with a ResNet-101 \cite{he2016deep} backbone). \\
\textbf{Implementation Details}:  We train our network on Titan
RTX GPU for 40K training iterations and a batch size of 2. We adopted the multi-resolution training strategy from HRDA \cite{hoyer2022hrda}. We adopted ADAMW \cite{loshchilov2017decoupled} optimizer with a learning rate of $6 \times 10^{-5}$ for encoder and $6 \times 10^{-4}$ for the decoder with linear learning rate warmup. The applicable strategies like DACS \cite{tranheden2021dacs} data augmentation, Rare Class Sampling \cite{hoyer2022daformer}, and ImageNet Feature Distance \cite{hoyer2022daformer} is used as it is along with corresponding set of parameters from the respective UDA methods. For EMA teacher update we have used $\alpha = 0.999$. Following the setting \cite{hoyer2022daformer, hoyer2022hrda} we also adopted color augmentation (brightness, contrast, saturation, hue, and blur) during source domain training. We used mean intersection-over-union (mIoU) as the metric to evaluate our UDA method.\\
\textbf{Reproducibility}: Our code is based on Pytorch \cite{imambi2021pytorch} and will be publicly available to reproduce all the results.
\begin{table}[]
\centering
\resizebox{\columnwidth}{!}{%
\begin{tabular}{lllll}
\toprule
Network   & UDA Method & w/o SegDA &  w/ SegDA & Diff. \\ \midrule
DeepLabV2 \cite{chen2017deeplab} & Entropy Min. \cite{vu2019advent}       &    44.3                &    49.2              &    +4.9            \\
DeepLabV2 \cite{chen2017deeplab} & DACS \cite{tranheden2021dacs}                &    53.9                &   56.5               &   +2.6             \\
DeepLabV2 \cite{chen2017deeplab} & DAFormer \cite{hoyer2022daformer}           &   56.0                &   59.8               &    +3.8           \\
DeepLabV2 \cite{chen2017deeplab} & HRDA \cite{hoyer2022hrda}               &     63.0               &   64.3               &     +1.3           \\ \midrule
DAFormer \cite{hoyer2022daformer}  & DAFormer \cite{hoyer2022daformer}            &    68.3               &   70.8               &   +2.5             \\
DAFormer \cite{hoyer2022daformer}  & HRDA \cite{hoyer2022hrda}               &    73.8               &    76.0              &    +2.2            \\ \bottomrule
\end{tabular}%
}
\caption{Performance (mIoU in \%) comparison of different UDA methods with and without SegDA on GTA $\rightarrow$ CS}
\label{tab:base_comp}
\end{table}
\begin{table*}[ht]
\centering
\resizebox{\textwidth}{!}{%
\begin{tabular}{lcccccccccccccccccccc}
\toprule
\multicolumn{1}{l|}{Method}     & \multicolumn{1}{l}{Road} & \multicolumn{1}{l}{S.walk} & \multicolumn{1}{l}{Build.} & \multicolumn{1}{l}{Wall} & \multicolumn{1}{l}{Fence} & \multicolumn{1}{l}{Pole} & \multicolumn{1}{l}{Tr.Light} & \multicolumn{1}{l}{Sign} & \multicolumn{1}{l}{Veget.} & \multicolumn{1}{l}{Terrain} & \multicolumn{1}{l}{Sky} & \multicolumn{1}{l}{Person} & \multicolumn{1}{l}{Rider} & \multicolumn{1}{l}{Car} & \multicolumn{1}{l}{Truck} & \multicolumn{1}{l}{Bus} & \multicolumn{1}{l}{Train} & \multicolumn{1}{l}{M.bike} & \multicolumn{1}{l|}{Bike} & \multicolumn{1}{l}{mIoU} \\ \midrule
\multicolumn{21}{c}{\textbf{Synthetic-to-Real: GTA→Cityscapes}}                                                                                                                                                                                                                                                                                                                                                                                                                                                                                                                                                \\ \midrule
\multicolumn{1}{l|}{ADVENT \cite{vu2019advent}}     & 89.4                     & 33.1                       & 81.0                       & 26.6                     & 26.8                      & 27.2                     & 33.5                         & 24.7                     & 83.9                       & 36.7                        & 78.8                    & 58.7                       & 30.5                      & 84.8                    & 38.5                      & 44.5                    & 1.7                       & 31.6                       & \multicolumn{1}{c|}{32.4} & 45.5                     \\
\multicolumn{1}{l|}{DACS \cite{tranheden2021dacs}}       & 89.9                     & 39.7                       & 87.9                       & 30.7                     & 39.5                      & 38.5                     & 46.4                         & 52.8                     & 88.0                       & 44.0                        & 88.8                    & 67.2                       & 35.8                      & 84.5                    & 45.7                      & 50.2                    & 0.0                       & 27.3                       & \multicolumn{1}{c|}{34.0} & 52.1                     \\
\multicolumn{1}{l|}{ProDA \cite{zhang2021prototypical}}      & 87.8                     & 56.0                       & 79.7                       & 46.3                     & 44.8                      & 45.6                     & 53.5                         & 53.5                     & 88.6                       & 45.2                        & 82.1                    & 70.7                       & 39.2                      & 88.8                    & 45.5                      & 59.4                    & 1.0                       & 48.9                       & \multicolumn{1}{c|}{56.4} & 57.5                     \\
\multicolumn{1}{l|}{DAFormer \cite{hoyer2022daformer}}   & 95.7                     & 70.2                       & 89.4                       & 53.5                     & 48.1                      & 49.6                     & 55.8                         & 59.4                     & 89.9                       & 47.9                        & 92.5                    & 72.2                       & 44.7                      & 92.3                    & 74.5                      & 78.2                    & 65.1                      & 55.9                       & \multicolumn{1}{c|}{61.8} & 68.3                     \\
\multicolumn{1}{l|}{HRDA \cite{hoyer2022hrda}}       & \underline{96.4}                     & \underline{74.4}                       & \underline{91.0}                       & \textbf{61.6}                     & \underline{51.5}                      & \underline{57.1}                     & \underline{63.9}                         & \underline{69.3}                     & \underline{91.3}                       & \underline{48.4}                        & \underline{94.2}                    & \underline{79.0}                       & \underline{52.9}                      & \underline{93.9}                    & \underline{84.1}                      & \underline{85.7}                    & \underline{75.9}                      & \underline{63.9}                       & \multicolumn{1}{c|}{\underline{67.5}} & \underline{73.8}                     \\
\multicolumn{1}{l|}{SegDA+ HRDA} & \multicolumn{1}{c}{\textbf{97.7}}     & \multicolumn{1}{c}{\textbf{80.1}}       & \multicolumn{1}{c}{\textbf{91.4}}       & \multicolumn{1}{c}{\textbf{61.6}}     & \multicolumn{1}{c}{\textbf{56.9}}      & \multicolumn{1}{c}{\textbf{59.8}}     & \multicolumn{1}{c}{\textbf{66.1}}         & \multicolumn{1}{c}{\textbf{71.4}}     & \multicolumn{1}{c}{\textbf{91.8}}       & \multicolumn{1}{c}{\textbf{51.6}}        & \multicolumn{1}{c}{\textbf{94.5}}    & \multicolumn{1}{c}{\textbf{79.9}}       & \multicolumn{1}{c}{\textbf{56.2}}      & \multicolumn{1}{c}{\textbf{94.7}}    & \multicolumn{1}{c}{\textbf{85.5}}      & \multicolumn{1}{c}{\textbf{90.4}}    & \multicolumn{1}{c}{\textbf{80.5}}      & \multicolumn{1}{c}{\textbf{64.5}}       & \multicolumn{1}{c|}{\textbf{68.5}}     & \multicolumn{1}{c}{\textbf{76.0}}     \\ \midrule
\multicolumn{21}{c}{\textbf{Synthetic-to-Real: Synthia→Cityscapes}}                                                                                                                                                                                                                                                                                                                                                                                                                                                                                                                                            \\ \midrule
\multicolumn{1}{l|}{ADVENT \cite{vu2019advent}}     & 85.6                     & 42.2                       & 79.7                       & 8.7                      & 0.4                       & 25.9                     & 5.4                          & 8.1                      & 80.4                       & –                           & 84.1                    & 57.9                       & 23.8                      & 73.3                    & –                         & 36.4                    & –                         & 14.2                       & \multicolumn{1}{c|}{33.0} & 41.2                     \\
\multicolumn{1}{l|}{DACS \cite{tranheden2021dacs}}       & 80.6                     & 25.1                       & 81.9                       & 21.5                     & 2.9                       & 37.2                     & 22.7                         & 24.0                     & 83.7                       & –                           & 90.8                    & 67.6                       & 38.3                      & 82.9                    & –                         & 38.9                    & –                         & 28.5                       & \multicolumn{1}{c|}{47.6} & 48.3                     \\
\multicolumn{1}{l|}{ProDA \cite{zhang2021prototypical}}      & \textbf{87.8}                     & 45.7                       & 84.6                       & 37.1                     & 0.6                       & 44.0                     & 54.6                         & 37.0                     & 88.1                       & –                           & 84.4                    & 74.2                       & 24.3                      & 88.2                    & –                         & 51.1                    & –                         & 40.5                       & \multicolumn{1}{c|}{45.6} & 55.5                     \\
\multicolumn{1}{l|}{DAFormer \cite{hoyer2022daformer}}   & 84.5                     & 40.7                       & 88.4                       & 41.5                     & 6.5                       & 50.0                     & 55.0                         & 54.6                     & 86.0                       & –                           & 89.8                    & 73.2                       & 48.2                      & 87.2                    & –                         & 53.2                    & –                         & 53.9                       & \multicolumn{1}{c|}{61.7} & 60.9                     \\
\multicolumn{1}{l|}{HRDA \cite{hoyer2022hrda}}       & 85.2                     & \underline{47.7}                       & \underline{88.8}                       & \underline{49.5}                     & 4.8                       & 57.2                     & 65.7                         & 60.9                     & 85.3                       & –                           & 92.9                    & 79.4                       & 52.8                      & 89.0                    & –                         & 64.7                    & –                         & 63.9                       & \multicolumn{1}{c|}{64.9} & 65.8                     \\
\multicolumn{1}{l|}{SegDA+ HRDA} & \multicolumn{1}{c}{87.2}     & \multicolumn{1}{c}{\textbf{50.7}}       & \multicolumn{1}{c}{\textbf{89.4}}       & \multicolumn{1}{c}{\textbf{49.6}}     & \multicolumn{1}{c}{\textbf{8.2}}      & \multicolumn{1}{c}{\textbf{59.6}}     & \multicolumn{1}{c}{\textbf{66.8}}         & \multicolumn{1}{c}{\textbf{63.6}}     & \multicolumn{1}{c}{\textbf{88.2}}       & \multicolumn{1}{c}{-}        & \multicolumn{1}{c}{\textbf{94.6}}    & \multicolumn{1}{c}{\textbf{81.0}}       & \multicolumn{1}{c}{\textbf{58.9}}      & \multicolumn{1}{c}{\textbf{90.2}}    & \multicolumn{1}{c}{-}      & \multicolumn{1}{c}{\textbf{64.7}}    & \multicolumn{1}{c}{-}      & \multicolumn{1}{c}{\textbf{67.1}}       & \multicolumn{1}{c|}{\textbf{64.9}}     & \multicolumn{1}{c}{\textbf{67.8}}     \\ \midrule
\multicolumn{21}{c}{\textbf{Day-to-Nighttime: Cityscapes→DarkZurich}}                                                                                                                                                                                                                                                                                                                                                                                                                                                                                                                                          \\ \midrule
\multicolumn{1}{l|}{ADVENT \cite{vu2019advent}}     & 85.8                     & 37.9                       & 55.5                       & 27.7                     & 14.5                      & 23.1                     & 14.0                         & 21.1                     & 32.1                       & 8.7                         & 2.0                     & 39.9                       & 16.6                      & 64.0                    & 13.8                      & 0.0                     & 58.8                      & 28.5                       & \multicolumn{1}{c|}{20.7} & 29.7                     \\
\multicolumn{1}{l|}{MGCDA \cite{sakaridis2020map}}      & 80.3                     & 49.3                       & 66.2                       & 7.8                      & 11.0                      & 41.4                     & 38.9                         & 39.0                     & 64.1                       & 18.0                        & 55.8                    & 52.1                       & 53.5                      & 74.7                    & 66.0                      & 0.0                     & 37.5                      & 29.1                       & \multicolumn{1}{c|}{22.7} & 42.5                     \\
\multicolumn{1}{l|}{DANNet \cite{wu2021dannet}}     & 90.0                     & 54.0                       & 74.8                       & 41.0                     & 21.1                      & 25.0                     & 26.8                         & 30.2                     & 72.0                       & 26.2                        & 84.0                    & 47.0                       & 33.9                      & 68.2                    & 19.0                      & 0.3                     & 66.4                      & 38.3                       & \multicolumn{1}{c|}{23.6} & 44.3                     \\
\multicolumn{1}{l|}{DAFormer \cite{hoyer2022daformer}}   & 93.5                     & 65.5                       & 73.3                       & 39.4                     & 19.2                      & 53.3                     & 44.1                         & 44.0                     & 59.5                       & 34.5                        & 66.6                    & 53.4                       & 52.7                      & 82.1                    & 52.7                      & 9.5                     & 89.3                      & 50.5                       & \multicolumn{1}{c|}{38.5} & 53.8                     \\
\multicolumn{1}{l|}{HRDA \cite{hoyer2022hrda}}       & 90.4                     & 56.3                       & 72.0                       & 39.5                     & 19.5                      & 57.8                     & 52.7                         & 43.1                     & 59.3                       & 29.1                        & 70.5                    & 60.0                       & 58.6                      & 84.0                    & 75.5                      & 11.2                    & 90.5                      & 51.6                       & \multicolumn{1}{c|}{40.9} & 55.9                     \\
\multicolumn{1}{l|}{SegDA+ HRDA} & \multicolumn{1}{c}{\textbf{94.8}}     & \multicolumn{1}{c}{\textbf{75.2}}       & \multicolumn{1}{c}{\textbf{84.1}}       & \multicolumn{1}{c}{\textbf{55.3}}     & \multicolumn{1}{c}{\textbf{28.7}}      & \multicolumn{1}{c}{\textbf{62.1}}     & \multicolumn{1}{c}{\textbf{52.7}}         & \multicolumn{1}{c}{\textbf{52.7}}     & \multicolumn{1}{c}{59.3}       & \multicolumn{1}{c}{\textbf{46.9}}        & \multicolumn{1}{c}{70.5}    & \multicolumn{1}{c}{\textbf{65.4}}       & \multicolumn{1}{c}{\textbf{61.8}}      & \multicolumn{1}{c}{\textbf{84.1}}    & \multicolumn{1}{c}{\textbf{75.6}}      & \multicolumn{1}{c}{\textbf{18.5}}    & \multicolumn{1}{c}{\textbf{91.3}}      & \multicolumn{1}{c}{\textbf{52.7}}       & \multicolumn{1}{c|}{ \textbf{44.3}}     & \multicolumn{1}{c}{\textbf{61.8}}     \\ \midrule
\multicolumn{21}{c}{\textbf{Clear-to-Adverse-Weather: Cityscapes→ACDC}}                                                                                                                                                                                                                                                                                                                                                                                                                                                                                                                                        \\ \midrule
\multicolumn{1}{l|}{ADVENT \cite{vu2019advent}}     & 72.9                     & 14.3                       & 40.5                       & 16.6                     & 21.2                      & 9.3                      & 17.4                         & 21.2                     & 63.8                       & 23.8                        & 18.3                    & 32.6                       & 19.5                      & 69.5                    & 36.2                      & 34.5                    & 46.2                      & 26.9                       & \multicolumn{1}{c|}{36.1} & 32.7                     \\
\multicolumn{1}{l|}{MGCDA \cite{sakaridis2020map}}      & 73.4                     & 28.7                       & 69.9                       & 19.3                     & 26.3                      & 36.8                     & 53.0                         & 53.3                     & 75.4                       & 32.0                        & 84.6                    & 51.0                       & 26.1                      & 77.6                    & 43.2                      & 45.9                    & 53.9                      & 32.7                       & \multicolumn{1}{c|}{41.5} & 48.7                     \\
\multicolumn{1}{l|}{DANNet \cite{wu2021dannet}}     & 84.3                     & 54.2                       & 77.6                       & 38.0                     & 30.0                      & 18.9                     & 41.6                         & 35.2                     & 71.3                       & 39.4                        & 86.6                    & 48.7                       & 29.2                      & 76.2                    & 41.6                      & 43.0                    & 58.6                      & 32.6                       & \multicolumn{1}{c|}{43.9} & 50.0                     \\
\multicolumn{1}{l|}{DAFormer \cite{hoyer2022daformer}}   & 58.4                     & 51.3                       & 84.0                       & 42.7                     & 35.1                      & 50.7                     & 30.0                         & 57.0                     & 74.8                       & 52.8                        & 51.3                    & 58.3                       & 32.6                      & 82.7                    & 58.3                      & 54.9                    & 82.4                      & 44.1                       & \multicolumn{1}{c|}{50.7} & 55.4                     \\
\multicolumn{1}{l|}{HRDA \cite{hoyer2022hrda}}       & 88.3                     & 57.9                       & 88.1                       & 55.2                     & 36.7                      & 56.3                     & 62.9                         & 65.3                     & 74.2                       & 57.7                        & 85.9                    & 68.8                       & 45.7                      & 88.5                    & 76.4                      & 82.4                    & 87.7                      & 52.7                       & \multicolumn{1}{c|}{60.4} & 68.0                     \\
\multicolumn{1}{l|}{SegDA+ HRDA} & \multicolumn{1}{c}{\textbf{90.8}}     & \multicolumn{1}{c}{ \textbf{67.4}}       & \multicolumn{1}{c}{ \textbf{89.3}}       & \multicolumn{1}{c}{\textbf{55.3}}     & \multicolumn{1}{c}{\textbf{40.5}}      & \multicolumn{1}{c}{\textbf{57.2}}     & \multicolumn{1}{c}{\textbf{62.9}}         & \multicolumn{1}{c}{\textbf{68.5}}     & \multicolumn{1}{c}{\textbf{76.4}}       & \multicolumn{1}{c}{\textbf{61.9}}        & \multicolumn{1}{c}{\textbf{87.1}}    & \multicolumn{1}{c}{ \textbf{71.4}}       & \multicolumn{1}{c}{ \textbf{49.5}}      & \multicolumn{1}{c}{\textbf{89.8}}    & \multicolumn{1}{c}{\textbf{76.5}}      & \multicolumn{1}{c}{\textbf{86.8}}    & \multicolumn{1}{c}{ \textbf{89.2}}      & \multicolumn{1}{c}{\textbf{56.9}}       & \multicolumn{1}{c|}{\textbf{63.3}}     & \multicolumn{1}{c}{\textbf{70.6}}     \\ \bottomrule
\end{tabular}%
}
\caption{Semantic Segmentation performance (mIoU in \%) for four UDA benchmarks}
\label{tab:comp}
\end{table*}

\begin{figure*}[htb]
\centering
\setlength\tabcolsep{0.2pt}
\begin{tabular}{ccccc}
 Image & Ground Truth & DAFormer\cite{hoyer2022daformer} & HRDA \cite{hoyer2022hrda} & SegDA + HRDA \\
 \animagetgt & \animagegt & \animageda & \animagehrda & \animageSegDA \\
\animagetgttwo & \animagegttwo& \animagedatwo & \animagehrdatwo & \animageSegDAtwo\\
 \animagetgtthree & \animagegtthree & \animagedathree & \animagehrdathree & \animageSegDAthree\\
  \animagetgtfour & \animagegtfour & \animagedafour & \animagehrdafour & \animageSegDAfour\\
\end{tabular}
\includegraphics[width=\textwidth]{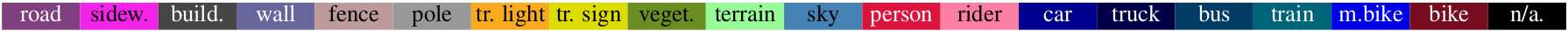}
  \caption{Qualitative comparison of SegDA with previous methods on GTA→CS (row 1 and 2), CS→ACDC (row 3), and CS→DarkZurich
(row 4)}
\label{fig:qual}
\end{figure*}

\subsection{Comparisons with State-of-the-art Methods}
To facilitate the comparison of SegDA with various SOTA methods, we first evaluated SegDA with different existing network architectures and UDA methods for domain adaptive semantic segmentation on GTA $\rightarrow$ CS. As shown in Table \ref{tab:base_comp}, with SegDA all the network architectures and UDA methods perform consistently better (ranging from +1.3 upto +4.9 mIoU) than without SegDA counterparts. This implies that the proposed domain adaptive framework SegDA not only benefit the CNN based architectures (like DeepLabV2 \cite{chen2017deeplab}) but able to perform better with Transformer based architectures as well like DAFormer \cite{hoyer2022daformer}. As expected the performance improvement with advanced transformer architectures is less since the base performance of UDA w/o SegDA is already high in these cases.

Going forward, we evaluate the performance of SegDA with the highest performing UDA method HRDA \cite{hoyer2022hrda} for further comparison with SOTA methods on different UDA scenarios namely: synthetic-to-real (GTA $\rightarrow$ CS and SYNTHIA $\rightarrow$ CS), clear-to-adverse weather (CS $\rightarrow$ ACDC) and day-to-nighttime (CS $\rightarrow$ DarkZurich). The quantitative comparison among different SOTA methods has been shown in Table \ref{tab:comp}, and the qualitative comparison is shown (in Figure \ref{fig:qual}) in the form of a visual difference between the image, ground truth, two latest SOTA transformer based methods \cite{hoyer2022daformer, hoyer2022hrda} and the proposed method SegDA. Summarizing results from Table \ref{tab:comp} SegDA outperforms both CNN based and Transformer based architectures including the recent transformer based SOTA methods namely, DAFormer \cite{hoyer2022daformer} and HRDA \cite{hoyer2022hrda}. It improves the state-of-the-art performance by +2.2 mIoU on GTA $\rightarrow$ Cityscapes(CS), +2.0 mIoU on SYNTHIA $\rightarrow$ Cityscapes, +5.9 mIoU on Cityscapes $\rightarrow$ DarkZurich, +2.6 mIoU on Cityscapes $\rightarrow$ ACDC. Moreover, SegDA performs better than SOTA on class-wise IoU as well on most of the classes. Specifically, it outperforms for all the classes in GTA $\rightarrow$ Cityscapes except on Wall which is a general class and mostly been occluded with different objects. Similarly, the performance per class on Cityscapes $\rightarrow$ ACDC outperforms the SOTA performance on each class, this proves the performance gain owning to ETF classifier and noise correction which handles the noise and separability even in the presence of adverse weather. Across UDA benchmarks, the classes that are on 
most advantage with SegDA are \textit{Fence}, \textit{Pole}, \textit{Traffic Light}, \textit{Terrain} and \textit{Rider}. ETF classifier and noise correction enforces the representation learned for these classes to be separable from commonly occurring classes, since they are dependent on context clues, HRDA models that better and ETF classifier helps to separate them from the most of the co-occurring classes. Most of the classes like \textit{Road}, \textit{Building} and \textit{Vegetation} are showing the least improvements since they are pretty much general and hence easiest to identify therefore their IoU scores are already high by SOTA methods and hence very little improvement. The classes like \textit{Bus} and \textit{Vegetation} are not discoverable satisfactorily on Cityscapes $\rightarrow$ DarkZurich because of the nighttime, it is not able to identify the green color for vegetation and color of vehicle as compared to black color which is present everywhere.\\
\textbf{Qualitative Comparison}: In Figure \ref{fig:qual}, the visual illustration of segmentation results shown to facilitate the comparison of the proposed method SegDA + HRDA over other two SOTA methods DAFormer\cite{hoyer2022daformer} and HRDA\cite{hoyer2022hrda} along with the corresponding ground truth for the image. Row 1 and 2 indicates segmentation results on GTA $\rightarrow$ CS, row 3 and 4 correspond to CS $\rightarrow$ ACDC and CS $\rightarrow$ DarkZurich respectively. The results highlighted by white dash boxes are the one captured correctly by SegDA as compared to other methods. Like n row 1, the SegDA is able to detect the \textit{car} correctly while HRDA highlight the \textit{car} along with \textit{traffic sign} (blue along with yellow color). Also, we see that SegDA is able to identify the \textit{traffic sign} correctly in row 2. All the segmentation results show the clear boundaries for the classes like \textit{pole}, \textit{fence}, \textit{sidewalk} etc. This is because of the ETF classifier making the pixel representation to be maximally separable and noise correction further enhances the consistency at the pixel level.
\begin{table}[ht]
\centering
\caption{Ablation Study of SegDA with DAFormer\cite{hoyer2022daformer} on GTA $\rightarrow$ CS}
\resizebox{\columnwidth}{!}{%
\begin{tabular}{lccccc}
\toprule
  & ETF Classifier & Color Aug & EMA Teacher & Noise Correction & mIoU \\ \midrule\midrule
1 & -              & -         & -           & -                &  68.3    \\
2 &    \checkmark            &     \checkmark       &    \checkmark          &    \checkmark               &  70.8   \\\midrule
3 &     -           &     \checkmark       &   \checkmark           &      \checkmark             &  68.6    \\
4 &       \checkmark         &       -    &      \checkmark        &     \checkmark              &  70.6    \\
5 &      \checkmark           &     \checkmark       &    -         &     \checkmark              &  69.4    \\
6 &           \checkmark      &        \checkmark    &     \checkmark         &  -                &  66.8    \\ \bottomrule
\end{tabular}%
}
\label{tab:ab}
\end{table}

\subsection{Ablation Studies}
In this section we present the analysis (in Table \ref{tab:ab}) of each of the individual components present in training of SegDA with DAFormer \cite{hoyer2022daformer} (due to faster training) on GTA $\rightarrow$ CS. Training SegDA end-to-end with DAFormer \cite{hoyer2022daformer}(row 2) achieves +2.5 mIoU better than DAFormer alone (row 1). Further ablations in row 3-6 remove one component at each row indicated by '-', first we remove the ETF Classifier structure instead used the single layer MLP in place of that this setting reduces the performance by -2.2 mIoU resulting in almost the same performance as with DAFormer as the segment representations obtained from Transformer is not adapted and hence the noise handling is itself noisy which leads to EMA teacher being noisy and hence no advantage. Ablation with color augmentation reduces the performance only by -0.2 mIoU which is insignificant and hence not as much important as ETF classifier. Other components like Noise Correction (along with discovery) and EMA Teacher shows the performance reduction of -4.0 mIoU and -1.4 mIoU respectively. Indicating the highest importance of Noise correction if ETF classifier is present without which the domain adaptation is not satisfactorily since it guides how to discover the pixels for new classes and correct the confidence score of existing pixel predictions. The EMA teacher ablation confirms the stability of pseudo labels. With latest model in place of EMA Teacher the resultant domain adaptation reduces by -1.4 mIoU because of fluctuating predictions of pixels from one model to another and hence averaging makes the prediction consistently confident.\\
\begin{table}[]
\centering
\caption{Relative comparison of UDA GTA $\rightarrow$ CS and Supervised Training on CS. Rel. indicates $\textrm{mIoU}_{UDA}/\textrm{mIoU}_{Superv.}$}
\small
\begin{tabular}{lccl}
\midrule
\textbf{}                & \textbf{$\textrm{mIoU}_{UDA}$} & \textbf{$\textrm{mIoU}_{Superv.}$} & \textbf{Rel.} \\ \midrule\midrule
DAFormer       & 68.3             & 77.6              &     88.0\%          \\
SegDA + DAFormer & 70.8          & 77.8                  &     91.0\%          \\ \midrule
Improvement     & +2.5              & +0.2                 &  +3.0\%             \\ \bottomrule
\end{tabular}
\label{tab:sup}
\end{table}
\textbf{Supervised Training}: We compared the supervised and UDA performance of DAFormer with and without SegDA in Table \ref{tab:sup}. For SegDA with DAFormer on supervised setting the still the pseudo labels are generated and using EMA teacher and noise corrected towards the loss correction along with segement representation adaptation. This setting leads to very little improvement of +0.2 mIoU over the DAFormer alone. For UDA however the imporvement is +2.5 mIoU indicating the usefulness of segment representation adaption, ETF classifier, noise correction and EMA Teacher for daomin adaptation and generating stable pseudo labels. To quantify this relative improvement of eaach network setting for UDA and Supervised training we calculated Rel. as $\textrm{mIoU}_{UDA}/\textrm{mIoU}_{Superv.}$ indicated in last column of Table \ref{tab:sup} indicating the DAFormer results in 88\% improvement on UDA and 91\% with SegDA applied over DAFormer. Overall there is 3\% improvement by an addition of SegDA over DAFormer.

\section{Conclusion}
We proposed the UDA method SegDA, which is able to maximally separate the visually correlated classes with the method of noise correction in the pseudo labels as well as pixel discovery to the classes not present in the pseudo labels. Our method outperforms (on existing SOTA) by +2.2 mIoU on GTA → Cityscapes, +2.0 mIoU on Synthia → Cityscapes, +5.9 mIoU on Cityscapes → DarkZurich, +2.6 mIoU on Cityscapes → ACDC.
\newpage
{\small
\bibliographystyle{ieee_fullname}
\bibliography{egpaper_final}
}

\end{document}